\documentclass[10pt,twocolumn,letterpaper]{article}
\usepackage[accsupp]{axessibility}  

\usepackage[pagenumbers]{cvpr} 

\usepackage[group-separator={,}]{siunitx} 

\usepackage{microtype}
\usepackage{caption}
\usepackage{booktabs}   
\usepackage{multirow}   
\usepackage{array}      
\renewcommand{\paragraph}[1]{\vspace{.3em}\noindent\textbf{#1.}}

\usepackage{soul}
\setuldepth{foobar}

\definecolor{cvprblue}{rgb}{0.21,0.49,0.74}
\usepackage[pagebackref,breaklinks,colorlinks,allcolors=cvprblue]{hyperref}

\def\paperID{} 
\def\confName{CVPR}
\def\confYear{2026}

\title{SubspaceAD: Training-Free Few-Shot Anomaly Detection \\ via Subspace Modeling}

\author{
Camile Lendering \quad
Erkut Akdag \quad
Egor Bondarev\\
AIMS Group, Department of Electrical Engineering, Eindhoven University of Technology\\
{\tt\small \{c.r.lendering, e.akdag, e.bondarev\}@tue.nl}
}

\begin{document}
\maketitle
\begin{abstract}
Detecting visual anomalies in industrial inspection often requires training with only a few normal images per category. Recent few-shot methods achieve strong results employing foundation-model features, but typically rely on memory banks, auxiliary datasets, or multi-modal tuning of vision-language models. We therefore question whether such complexity is necessary given the feature representations of vision foundation models. To answer this question, we introduce \textbf{SubspaceAD}, a training-free method, that operates in two simple stages. First, patch-level features are extracted from a small set of normal images by a frozen DINOv2 backbone. Second, a Principal Component Analysis (PCA) model is fit to these features to estimate the low-dimensional subspace of normal variations. At inference, anomalies are detected via the reconstruction residual with respect to this subspace, producing interpretable and statistically grounded anomaly scores. Despite its simplicity, SubspaceAD achieves state-of-the-art performance across one-shot and few-shot settings without training, prompt tuning, or memory banks. In the one-shot anomaly detection setting, SubspaceAD achieves image-level and pixel-level AUROC of 97.1\% and 97.5\% on the MVTec-AD dataset, and 93.2\% and 98.2\% on the VisA dataset, respectively, surpassing prior state-of-the-art results. Code and demo are available at~\url{https://github.com/CLendering/SubspaceAD}
\end{abstract}    
\section{Introduction}
\label{sec:intro}
Detecting visual anomalies in images is a long-standing challenge in computer vision~\cite{chandola2009anomaly,pang2021deep}. In industrial inspection, even subtle deviations from normal appearance, such as scratches, contaminations, or missing components, can lead to downstream failures or safety risks. Development of systems that automatically detect such defects is therefore essential for reliable and cost-efficient production.

The primary challenge in industrial anomaly detection (AD) is data scarcity: full-shot methods require hundreds of defect-free images per category to model normality, which is rarely feasible in practice. At the other extreme, zero-shot methods~\cite{jeong2023winclip, zhou2023anomalyclip, xu2024customizing} leverage vision-language models (VLMs) and textual prompts to detect anomalies without any normal samples. However, such methods often struggle with detection of subtle, non-semantic defects \textit{(e.g., small cracks)} that cannot be easily captured by language alone. This paper focuses on the practical and challenging \textit{few-shot} regime, where only a small number of normal images are available to define what constitutes normal appearance for a given object category.

\begin{figure}[t]
\centering
\includegraphics[width=\columnwidth]{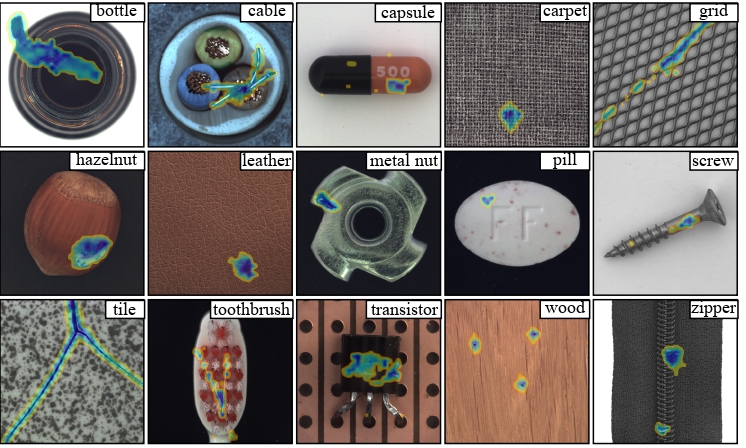}
\caption{One-shot segmentation results of \textit{SubspaceAD} on the MVTec-AD dataset~\cite{bergmann2019mvtec}, where \textit{SubspaceAD} only uses one normal image per category. Each example shows a test sample with its predicted anomaly mask (overlaid in dark blue), across all 15 categories of the MVTec-AD dataset.}
\label{fig:intro_figure}
\end{figure}

To address the few-shot challenge, recent studies have introduced increasingly complex deep learning techniques, which can be divided into three categories. The first one comprises reconstruction-based approaches~\cite{bergmann2018improving,schlegl2019f,fuvcka2024transfusion}, which learn to reproduce only normal samples and employ reconstruction residuals as indicators of abnormality. The second category relies on large memory banks of features~\cite{roth2022patchcore,damm2025anomalydino,cohen2020sub}, storing large collections of patch embeddings from normal images and performing anomaly detection via nearest-neighbor retrieval in feature space. More recently, VLM approaches have adapted models like CLIP~\cite{radford2021learning} through prompt tuning~\cite{jeong2023winclip,li2024promptad,lvone} to enable text-guided anomaly detection.

While methods across these three categories achieve strong performance on benchmarks such as MVTec-AD~\cite{bergmann2019mvtec} and VisA~\cite{zou2022spotvisa}, they have become increasingly complex. These methods often require extensive data augmentation, careful hyperparameter tuning, multi-stage training, auxiliary learning objectives, or large-footprint memory banks, making them difficult to deploy and maintain in real-world industrial settings. In parallel, representation learning has advanced substantially. Foundation vision models, such as DINOv2, produce dense and transferable features that capture both semantic and structural properties of images, even for domains they were never trained on~\cite{caron2021emerging,oquab2023dinov2,simeoni2025dinov3}. With features of such quality, one may ask a simple question: \textit{do we still need complex pipelines, large memory banks, and multi-stage tuning to detect anomalies?}

This paper argues that the answer is no. By leveraging strong foundation features, we show that a far simpler alternative is not only feasible but superior. Specifically, we propose a purely statistical approach based on Principal Component Analysis (PCA)~\cite{mackiewicz1993principal,pearson1901liii}. Given only a few normal images, PCA defines a low-dimensional subspace that captures the `principal' variation of normal appearance. Deviations from this subspace, quantified by the reconstruction residuals, directly indicate anomalies. This approach follows the well-established statistical principle: anomalies (or outliers) manifest as deviations from the dominant PCA subspace of normal data~\cite{shyu2003novel}.

This minimalist method, termed \textit{SubspaceAD}, is training-free, parameter-light, and interpretable. As demonstrated in Fig.~\ref{fig:intro_figure}, this simple formulation is powerful enough to localize diverse defect patterns even when provided with only a single normal reference sample per category. Extensive experiments show that SubspaceAD surpasses the performance of recently proposed reconstruction-based, memory-bank-based, and VLM-based approaches, suggesting that with sufficiently expressive features, classical statistical modeling can once again serve as a powerful foundation for visual anomaly detection. Summarizing, this paper provides the following contributions:
\begin{itemize}
    \item We introduce \textbf{SubspaceAD}, a minimalist, training-free method for few-shot ($k \in \{1,2,4\}$) anomaly detection that combines frozen DINOv2 features with PCA to model normal appearance.
    \item Through comprehensive evaluations on MVTec-AD and VisA datasets, SubspaceAD outperforms state-of-the-art reconstruction-, memory-bank-, and VLM-based approaches across all few-shot settings.
    \item SubspaceAD is interpretable and parameter-free, requiring only a single normal image per category and a single forward pass per test image.
\end{itemize}

\section{Related Work}
\label{sec:relatedworks}

\subsection{Reconstruction-Based Approaches}
Reconstruction-based approaches detect anomalies by learning to reproduce only normal samples and identifying deviations through reconstruction error. Early methods rely on autoencoders or variational autoencoders to reconstruct normal appearance, assuming that anomalies cannot be accurately recovered~\cite{bergmann2018improving,schlegl2019f}. Generative models extend this idea by aligning reconstructions with a learned normal data manifold as demonstrated in~\cite{DBLP:journals/corr/SchleglSWSL17, akcay2018ganomaly}. More recent developments introduce perceptual losses, diffusion-based priors, or feature regression strategies to avoid the common pitfall of over-generalization, where the model inadvertently learns to reconstruct anomalous patterns~\cite{fuvcka2024transfusion, deng2022anomaly}. One of the latest works, FastRecon~\cite{fang2023fastrecon} learns a transform matrix from a few normal samples to reconstruct features as normal, by regression with distribution regularization. While these methods have shown success across industrial defect benchmarks, they require explicit training, hyperparameter tuning, and careful balancing between reconstruction quality and anomaly sensitivity.

\subsection{Memory Bank-Based Anomaly Detection}
Another major direction in anomaly detection involves storing representative patch features of normal samples in a memory bank and identifying anomalies via nearest-neighbor matching. For instance, SPADE~\cite{cohen2020sub} uses multi-resolution feature correspondences inspired by $k$-NN and operates in a training-free manner, making it suitable for detecting anomalies in few-shot settings. PatchCore~\cite{roth2022patchcore}, another training-free anomaly detection method, reduces memory redundancy by selecting a compact core set of embeddings to improve retrieval efficiency. This method has demonstrated the ability to handle few-shot anomaly detection. Related approaches estimate feature distributions at spatial locations~\cite{defard2021padim}, use flow-based transformations for density modeling~\cite{yu2021fastflow}, or distill pre-trained teacher networks to compress normality priors~\cite{deng2022anomaly}. More recent methods such as AnomalyDINO~\cite{damm2025anomalydino} leverage features from vision foundation models (e.g., DINOv2) to improve both robustness and localization quality. Despite their strong performance, memory-bank methods typically require storing thousands to millions of patch descriptors and performing nearest-neighbor search at inference, which can become computationally heavy, especially in few-shot or multi-category deployment scenarios.

\subsection{Foundational Models and VLM-Based Approaches}
Large-scale foundation models~\cite{radford2021learning, li2024multimodal, liu2024grounding, he2022masked, chen2020simple}, including vision-only approaches such as, DINO~\cite{caron2021emerging,oquab2023dinov2}, have significantly influenced visual representation learning, both in the unimodal and multimodal domains. With the success of large-scale vision-language models like CLIP~\cite{radford2021learning}, recent works have explored leveraging text prompts for anomaly detection. For instance, WinCLIP~\cite{jeong2023winclip} is one of the first works to adopt CLIP for anomaly detection. It utilizes manually designed text prompts to detect anomalies across predefined multi-scale windows, while constructing a multi-scale memory bank for feature matching in few-shot settings. Subsequent methods, like AnoVL~\cite{deng2023anovl} and PromptAD~\cite{li2024promptad}, automate prompt creation or learn prompt adapters, while others attempt to learn generic normality and abnormality prompts across categories using auxiliary datasets~\cite{zhou2023anomalyclip,cao2024adaclip}.
Specifically, PromptAD~\cite{li2024promptad} proposes semantic concatenation to reverse prompt’s semantics and directly optimize a set of learnable context vectors. IIPAD~\cite{lvone} instead generates prompts directly from the available normal instances rather than learning category-specific prompts. This enables a single shared prompt space that generalizes across categories, improving few-shot anomaly detection efficiency without extra training data. Although these approaches improve flexibility, they follow a one-class-per-prompt paradigm and often depend on additional normal/anomalous data, prompt tuning, or domain-specific textual priors.

\subsection{Few-Shot and Training-Free Anomaly Detection}
Few-shot anomaly detection methods vary in how they characterize normal variation. Training-free vision-based approaches, such as DN2~\cite{bergman2020deep}, SPADE~\cite{cohen2020sub} and PatchCore~\cite{santos2023optimizing} typically store normal patch features and detect anomalies via nearest-neighbor retrieval. Methods requiring fine-tuning, including PaDiM~\cite{defard2021padim} and GraphCore~\cite{xie2023pushing}, instead learn parametric models of feature distributions. Beyond purely visual pipelines, vision–language approaches such as ADP~\cite{kwak2024few}, WinCLIP~\cite{jeong2023winclip} and GPT-4V-based anomaly reasoning~\cite{xu2024customizing}, use text prompts or language alignment to guide few-shot detection, while zero-, few-shot like APRIL-GAN~\cite{chen2023zero} and AnomalyCLIP~\cite{zhou2023anomalyclip} aim to generalize across categories without additional training. Batched zero-shot frameworks  MuSc~\cite{li2024musc} and ACR~\cite{li2023zero} further exploit collective statistics across test sets rather than evaluating samples independently. Collectively, these approaches demonstrate a shift toward reducing supervision and eliminating training overhead, while maintaining strong anomaly discrimination. Our work follows this direction but departs from reliance on memory banks or prompt tuning by modeling normal variation through a simple PCA-based subspace formulation.
\section{Method}
\label{sec:methods}
\begin{figure*}[t]
\centering
\includegraphics[width=\linewidth,]{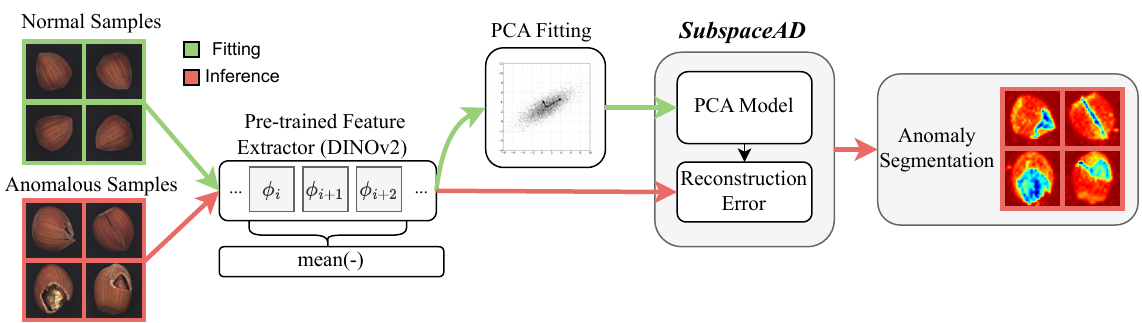}
\caption{{Overview of SubspaceAD. \textit{(Fitting)}:  Aggregated patch features are collected from $k$ normal samples using a frozen DINOv2-G model and a PCA model is fitted to capture the subspace of normal variation. \textit{(Inference)}: Features of a test image are extracted, projected onto the normal subspace, and the reconstruction error is computed, providing the anomaly segmentation map directly. \small{PCA figure from~\cite{pca_wiki}.}}}
\label{fig:method_overview}
\end{figure*}
The proposed \textit{SubspaceAD} method models the linear subspace of normal patch features, eliminating the need for memory banks, prompt tuning, or external data. Anomaly scores are computed via reconstruction errors from this subspace, resulting in a training-free, compact, and interpretable method. An overview of \textit{SubspaceAD} is provided in Fig.~\ref{fig:method_overview}, which operates in two straightforward stages. First, patch-level features are extracted from a small set of $k$ normal images by a frozen DINOv2-G backbone. Second, a Principal Component Analysis (PCA) model is fit to these features to estimate the low-dimensional subspace of normal variation. At inference, anomalies are detected and localized from the reconstruction residuals of test features with respect to this subspace. 

\subsection{Problem Formulation}
Given a small set of $k$ anomaly-free training images 
$\mathcal{I}_{\mathrm{train}} = \{I_1, \dots, I_k\}$ 
and a test image $I_{\mathrm{test}}$, 
the goal is to define an anomaly scoring function $A$, which predicts an anomaly likelihood for every spatial position $p$ in $I_{\mathrm{test}}$:
\begin{equation}
    A(I_{\mathrm{test}}, p) \in[0,1].
\end{equation}
In the few-shot regime, the key challenge is to model the manifold of normal patch features using only a limited number of clean samples. 
It is assumed that patch-level features from normal samples lie near a low-dimensional linear subspace embedded in the feature space of a foundation model, while anomalous regions correspond to samples with large reconstruction residuals outside this subspace.

\subsection{Feature Extraction}
\label{subsec:feature_extraction}

The core of SubspaceAD is the dense feature representation extracted from a pre-trained vision model. A frozen DINOv2-G model~\cite{oquab2023dinov2} is employed as the feature extraction model to obtain patch-level features. Given an input image, the model produces a sequence of patch tokens, where each token corresponds to a $14\times14$ patch of the image.

Crucially, instead of using only the tokens from the final transformer block, tokens are aggregated from multiple intermediate layers to obtain a more robust representation that balances high-level semantics with low-level spatial detail. This multi-layer fusion improves sensitivity to subtle anomalies while preserving global context cues, a design choice supported in the ablation study (Sec.~\ref{subsec:ablations}, Table~\ref{tab:ablation_layers}).

Let $f_l(p) \in \mathbb{R}^{D}$ denote the patch token at spatial position $p$ from transformer block $l$, where $D$ is the model's feature dimension (e.g., 1536 for DINOv2-G). 
Tokens are extracted from a set of layers $\mathcal{L}$ (layers 22--28 in DINOv2-G). The final feature vector $x_p \in \mathbb{R}^{D}$ for position $p$ is defined as the mean-pooled representation (multi-layer averaging):
\begin{equation}
    x_p = \frac{1}{|\mathcal{L}|} \sum_{l \in \mathcal{L}} f_l(p).
\end{equation}
This process yields a feature vector $x_p$ for each patch. Together, these vectors form a dense feature map $X \in \mathbb{R}^{h \times w \times D}$ for each image,  where $h$ and $w$ denote the spatial grid dimensions.

For PCA-based modeling, averaging the features across multiple intermediate layers is particularly beneficial. Since the anomaly score is derived from the residual variance orthogonal to the principal subspace, its reliability depends on how well the feature distribution captures meaningful structure rather than noise. Intermediate layers in DINOv2 contain a mix of semantic and structural information, whereas the deepest layers tend to collapse local detail into category-level abstractions. Therefore, averaging the features across several middle layers stabilizes the covariance estimation, reduces layer-specific variance, and ensures that the principal components represent stable patterns of normal appearance.

To build a representative covariance matrix from only $k$ normal images, data augmentation is applied.
For each of the $k$ normal images, $N_a=30$ augmented views are generated by applying random rotations between $0^{\circ}$ and $345^{\circ}$, as rotational variance is common in industrial inspection. Features are extracted from all $k \times (1 + N_a)$ images (the original, plus its augmentations), forming the set of all patch features as $X_{\mathrm{normal}}$. This ensures that the estimated subspace captures common geometric variations and is not biased by a single view. The method consists of a single fitting phase (on $k$ normal images) and an inference phase applied per test image, as illustrated in Fig.~\ref{fig:method_overview}.

\subsection{Subspace Modeling of Normal Features}
\label{subsec:subspace_modeling}
Normal patch features are modeled via Principal Component Analysis (PCA). This model is fit to the set of all normal features $X_{\mathrm{normal}}$, which contains all patch vectors $x_p$ collected from the $k$ original and augmented normal images (as defined in Sec.~\ref{subsec:feature_extraction}). From this set, the empirical mean $\mu \in \mathbb{R}^D$ and covariance matrix $\Sigma \in \mathbb{R}^{D \times D}$ are computed.

PCA gives a closed-form parameter-free estimate of the dominant linear subspace of the data. We use deterministic PCA for simplicity and numerical stability, making it well-suited for the proposed few-shot regime where overfitting must be avoided. Each patch feature $x \in \mathbb{R}^D$ is modeled as:
\begin{equation}
  x = \mu + C z + \epsilon,
 \quad z \sim \mathcal{N}(0, I_r),
 \quad \epsilon \sim \mathcal{N}(0, \sigma^2 I_C),
\end{equation}
where $C \in \mathbb{R}^{D \times r}$ contains the top $r$ eigenvectors of $\Sigma$, $z \in \mathbb{R}^r$ is a latent variable (with $I_r$ being the $r \times r$ identity matrix), and $\epsilon$ is an isotropic noise term (with $I_C$ being the $D \times D$ identity matrix). The matrix $C$ forms an orthonormal basis for the subspace of normal variation. Under this probabilistic formulation~\cite{tipping1999probabilistic}, the squared reconstruction residual  $\lVert (x - \mu) - CC^\top (x - \mu) \rVert_2^2$ corresponds to the negative log-likelihood component orthogonal to the subspace, thus defining an anomaly score.

The number of retained components $r$ is chosen such that the explained variance exceeds a predefined threshold $\tau$:
\begin{equation}
 \sum_{i=1}^r \lambda_i \;\ge\; \tau \sum_{i=1}^D \lambda_i,
 \qquad \tau = 0.99,
\end{equation}
where $\lambda_i$ denotes the $i$-th eigenvalue of $\Sigma$. This high threshold is chosen to ensure the subspace captures the vast majority of normal variation, while discarding minor noise components (see Sec.~\ref{subsec:ablations} for an empirical analysis). The resulting model is fully described by the mean vector $\mu$ and the basis matrix $C$.

\subsection{Anomaly Scoring and Localization}
\label{subsec:scoring}
For a test image, its corresponding patch feature map $X_{\mathrm{test}} \in \mathbb{R}^{h \times w \times D}$ is extracted as described in Sec.~\ref{subsec:feature_extraction}. This map consists of all patch feature vectors $x_p$ for the image.

\paragraph{Patch-level Scoring} 
Each patch feature vector $x_p$ is projected onto the normal subspace:
\begin{equation}
    x_{\mathrm{proj}} = \mu + CC^\top (x_p - \mu),
\end{equation}
and assigned a residual-based anomaly score given by:
\begin{equation}
    S(x_p) = \lVert x_p - x_{\mathrm{proj}} \rVert_2^2.
\end{equation}
This score measures the deviation of each feature vector from the principal subspace of normal variation, resulting in a low-resolution anomaly map $M \in \mathbb{R}^{h \times w}$. 

\paragraph{Image-level Aggregation} To aggregate patch-level scores into a single image-level prediction, we employ a tail-robust statistic, the empirical tail value-at-risk (TVaR), which averages the top $\rho\%$ of patch scores in the anomaly map $M$. Let $H_{\rho}(M)$ denote the set of scores in $M$ at or above the $(100-\rho)$-th percentile. The image-level score $s_{\mathrm{img}}$ is then computed as the mean of this set:
\begin{equation}
    s_{\mathrm{img}} = \operatorname{mean}\!\bigl(H_{\rho}(M)\bigr).
\end{equation}
We set $\rho = 1\%$, following prior work~\cite{damm2025anomalydino}, which balances sensitivity to subtle defects with robustness to sparse false positives.

\paragraph{Pixel-level Localization}
For visualization and pixel-level evaluation, the patch-level anomaly map $M$ is bilinearly upsampled to the original image resolution and smoothed using a Gaussian filter with $\sigma=4$ to suppress high-frequency noise while preserving localization accuracy.
Finally, the normalized anomaly score function is defined as
\begin{equation}
A(I_{\mathrm{test}}, p) = \mathrm{Norm}\big(S(x_p)\big),
\end{equation}
where $\mathrm{Norm}(\cdot)$ denotes min–max normalization to $[0,1]$. These normalized maps are used for visualization, while AUROC and PRO metrics are computed using the raw (unnormalized) scores.

\subsection{Complexity and Memory Analysis}
Let $n = k \times (1+ N_a) \times (h \times w)$ denote the total number of normal patch features (with $k$ normal training images) and $D$ the feature dimension. PCA fitting requires $O(nD^2)$ time for covariance computation and $O(D^3)$ for eigendecomposition, both negligible in the few-shot regime. The resulting model consists only of $\mu \in \mathbb{R}^D$ and $C \in \mathbb{R}^{D \times r}$, typically requiring less than 1~MB of storage per category.
Inference on a $672\times672$ image takes approximately 300~\text{ms} on a single NVIDIA H100 GPU, with the majority of time dominated by the DINOv2-G forward pass ($\sim$226~\text{ms}), and the subspace projection and the scoring take only $\sim$74~\text{ms} (see Appendix~\ref{sec:inference_time} for a hardware-normalized analysis). 

\section{Experiments}
\label{sec:experiments}

\begin{table*}[ht!]
\centering
\caption{Comparison of anomaly detection and localization performance on MVTec-AD and VisA across different few-shot settings. Results for SubspaceAD (ours) are reported as mean $\pm$ standard deviation over 5 seeds. Best results are in \textbf{bold}, and second-best are \underline{underlined}.}
\label{tab:main_comparison}
\small
\begin{tabular}{@{}llcccccccc@{}}
\toprule
 &  & \multicolumn{4}{c}{\textbf{MVTec-AD}} & \multicolumn{4}{c}{\textbf{VisA}} \\
\cmidrule(lr){3-6} \cmidrule(lr){7-10}
 &  & \multicolumn{2}{c}{Image-level} & \multicolumn{2}{c}{Pixel-level} & \multicolumn{2}{c}{Image-level} & \multicolumn{2}{c}{Pixel-level} \\
\cmidrule(lr){3-4} \cmidrule(lr){5-6} \cmidrule(lr){7-8} \cmidrule(lr){9-10}
\textbf{Setup} & \textbf{Method} & AUROC & AUPR & AUROC & PRO & AUROC & AUPR & AUROC & PRO \\
\midrule
0-shot & WinCLIP & 91.8 & 96.5 & 85.1 & 64.6 & 78.1 & 81.2 & 79.6 & 56.8 \\
\midrule
\multirow{8}{*}{1-shot} & SPADE & 81.0 & 90.6 & 91.2 & 83.9 & 79.5 & 82.0 & 95.6 & 84.1 \\
 & PatchCore & 83.4 & 92.2 & 92.0 & 79.7 & 79.9 & 82.8 & 95.4 & 80.5 \\
 & FastRecon & - & - & - & - & - & - & - & - \\
 & WinCLIP & 93.1 & 96.5 & 95.2 & 87.1 & 83.8 & 85.1 & 96.4 & 85.1 \\
 & PromptAD & 94.6 & 97.1 & 95.9 & 87.9 & 86.9 & 88.4 & 96.7 & 85.1 \\
 & IIPAD & 94.2 & 97.2 & 96.4 & 89.8 & 85.4 & 87.5 & 96.9 & 87.3 \\
 & AnomalyDINO & \underline{96.6} & \underline{98.2} & \underline{96.8} & \underline{92.7} & \underline{87.4} & \underline{89.0} & \underline{97.8} & \underline{92.5} \\
 & SubspaceAD (ours) & \textbf{97.1 $\pm$ 0.9} & \textbf{98.6 $\pm$ 0.4} & \textbf{97.5 $\pm$ 0.1} & \textbf{94.5 $\pm$ 0.2} & \textbf{93.2 $\pm$ 0.8} & \textbf{93.0 $\pm$ 0.8} & \textbf{98.2 $\pm$ 0.1} & \textbf{95.5 $\pm$ 0.1} \\
\midrule
\multirow{8}{*}{2-shot} & SPADE & 82.9 & 91.7 & 92.0 & 85.7 & 80.7 & 82.3 & 96.2 & 85.7 \\
 & PatchCore & 86.3 & 93.8 & 93.3 & 82.3 & 81.6 & 84.8 & 96.1 & 82.6 \\
 & FastRecon & 91.0 & - & 95.9 & - & - & - & - & - \\
 & WinCLIP & 94.4 & 97.0 & 96.0 & 88.4 & 84.6 & 85.8 & 96.8 & 86.2 \\
 & PromptAD & 95.7 & 97.9 & 96.2 & 88.5 & 88.3 & 90.0 & 97.1 & 85.8 \\
 & IIPAD & 95.7 & 97.9 & 96.7 & 90.3 & 86.7 & 88.6 & 97.2 & 87.9 \\
 & AnomalyDINO & \underline{96.9} & \underline{98.2} & \underline{97.0} & \underline{93.1} & \underline{89.7} & \underline{90.7} & \underline{98.0} & \underline{93.4} \\
 & SubspaceAD (ours) & \textbf{97.5 $\pm$ 0.7} & \textbf{98.8 $\pm$ 0.4} & \textbf{97.8 $\pm$ 0.1} & \textbf{94.9 $\pm$ 0.1} & \textbf{93.8 $\pm$ 0.4} & \textbf{93.4 $\pm$ 0.7} & \textbf{98.3 $\pm$ 0.0} & \textbf{95.7 $\pm$ 0.1} \\
\midrule
\multirow{8}{*}{4-shot} & SPADE & 84.8 & 92.5 & 92.7 & 87.0 & 81.7 & 83.4 & 96.6 & 87.3 \\
 & PatchCore & 88.8 & 94.5 & 94.3 & 84.3 & 85.3 & 87.5 & 96.8 & 84.9 \\
 & FastRecon & 94.2 & - & 97.0 & - & - & - & - & - \\
 & WinCLIP & 95.2 & 97.3 & 96.2 & 89.0 & 87.3 & 88.8 & 97.2 & 87.6 \\
 & PromptAD & 96.6 & 98.5 & 96.5 & 90.5 & 89.1 & 90.8 & 97.4 & 86.2 \\
 & IIPAD & 96.1 & 98.1 & 97.0 & 91.2 & 88.3 & 89.6 & 97.4 & 88.3 \\
 & AnomalyDINO & \underline{97.7} & \underline{98.7} & \underline{97.2} & \underline{93.4} & \underline{92.6} & \underline{92.9} & \underline{98.2} & \underline{94.1} \\
 & SubspaceAD (ours) & \textbf{98.0 $\pm$ 0.4} & \textbf{99.0 $\pm$ 0.2} & \textbf{97.9 $\pm$ 0.1} & \textbf{95.1 $\pm$ 0.1} & \textbf{94.7 $\pm$ 0.2} & \textbf{94.3 $\pm$ 0.5} & \textbf{98.4 $\pm$ 0.0} & \textbf{96.0 $\pm$ 0.1} \\
\bottomrule
\end{tabular}
\end{table*}
The performance of SubspaceAD is evaluated against recent state-of-the-art methods under 1-, 2-, and 4-shot settings, reporting both image-level and pixel-level results (Sec.~\ref{subsec:sota}). We further assess its generalization under the \textit{batched 0-shot} setting, where the entire unlabeled test set is modeled jointly without any reference images (Sec.~\ref{subsec:batched_0shot}). Finally, ablation studies are conducted to validate the model design choices, including the foundation-model backbone, input image resolution, layer aggregation strategy, and PCA explained-variance threshold~$\tau$ (Sec.~\ref{subsec:ablations}).

\subsection{Datasets}
\label{subsec:datasets}
SubspaceAD is evaluated on two widely used industrial anomaly detection benchmarks: MVTec-AD~\cite{bergmann2019mvtec} and VisA~\cite{zou2022spotvisa}. Both datasets contain multiple subsets of distinct object and texture categories. MVTec-AD contains 15 categories with image resolutions ranging from 700$\times$700 to 1024$\times$1024 pixels, while VisA includes higher-resolution images (around 1500$\times$1000 pixels) and a broader range of complex real-world anomaly types. Since anomaly detection is formulated as a one-class problem, the training set for each category consists only of normal (defect-free) samples, while the test set contains both normal and anomalous instances. Anomalies in the test set are annotated with ground-truth labels at both the image and pixel levels.

\subsection{Evaluation Metrics}
\label{subsec:evaluationmetrics}
Following standard practice~\cite{roth2022patchcore, damm2025anomalydino}, performance is evaluated at both the image and pixel levels. Image-level anomaly detection is measured by the Area Under the Receiver Operating Characteristic (AUROC) and Average Precision (AUPR). Pixel-level localization is assessed via pixel-wise AUROC and the Per-Region Overlap (PRO), which accounts for the spatial extent of anomalous regions.

\subsection{Implementation Details}
\label{subsec:implementationdetails}
The frozen DINOv2-G model~\cite{oquab2023dinov2} is deployed for feature extraction, with features averaged across layers 22--28, as described in Sec.~\ref{subsec:feature_extraction}. For few-shot fitting, $k \in \{1, 2, 4\}$ normal images are randomly selected, and $N_a=30$ random rotations (from $0^{\circ}$ to $345^{\circ}$) are applied to each, except for the orientation-sensitive \textit{transistor} category. For completeness, a fully rotation-agnostic evaluation is provided in Appendix~\ref{sec:rotation_agnostic}. The PCA variance threshold is set to $\tau=0.99$, and TVaR aggregation uses $\rho=1\%$~\cite{damm2025anomalydino}. Importantly, we apply a single, fixed image resolution of 672~px across all categories and shots for both MVTec-AD and VisA datasets (see Sec.~\ref{subsec:ablations} for sensitivity analysis). Each few-shot configuration is evaluated over 5 independent runs with different random seeds, reporting the mean and standard deviation. All experiments are performed on a single NVIDIA H100 GPU.

\subsection{Baselines}
\label{subsec:baselines}
We compare SubspaceAD against representative few-shot methods across three dominant paradigms: (1) \textit{memory-bank-based} approaches, including SPADE~\cite{cohen2020sub}, PatchCore~\cite{roth2022patchcore}, and AnomalyDINO~\cite{damm2025anomalydino}; (2) \textit{reconstruction-based} methods, such as FastRecon~\cite{fang2023fastrecon}; and (3) \textit{VLM-based} models, including WinCLIP~\cite{jeong2023winclip}, PromptAD~\cite{li2024promptad}, and IIPAD~\cite{lvone}.

\subsection{Comparison to the State-of-the-Art}
\label{subsec:sota}
Table \ref{tab:main_comparison} compares SubspaceAD with recent few-shot methods on MVTec-AD and VisA under 1-, 2-, and 4-shot settings. Using a 672 px resolution (Sec. \ref{subsec:ablations}), SubspaceAD consistently achieves new state-of-the-art average performance across all evaluated image- and pixel-level metrics.

On MVTec-AD, SubspaceAD attains 97.1\% image-level and 97.5\% pixel-level AUROC in the 1-shot setting, outperforming all prior methods. On the more challenging VisA benchmark, SubspaceAD achieves 93.2\% image-level AUROC and 95.5\% PRO, surpassing the prior state-of-the-art (AnomalyDINO) by 5.8\% and 3.0\%, respectively.

As the number of reference samples increases, our method maintains its lead across all metrics. In the 4-shot setting, SubspaceAD achieves the highest PRO on MVTec-AD (95.1\%) and VisA (96.0\%). These results validate our central claim: leveraging strong foundation features with a parameter-free statistical model can outperform more complex state-of-the-art methods. We also report performance under the full-shot setting (see Appendix~\ref{sec:full_shot}).

\begin{figure*}[t]
\centering
\begin{minipage}[t]{0.49\textwidth}
    \centering
    \includegraphics[width=\linewidth]{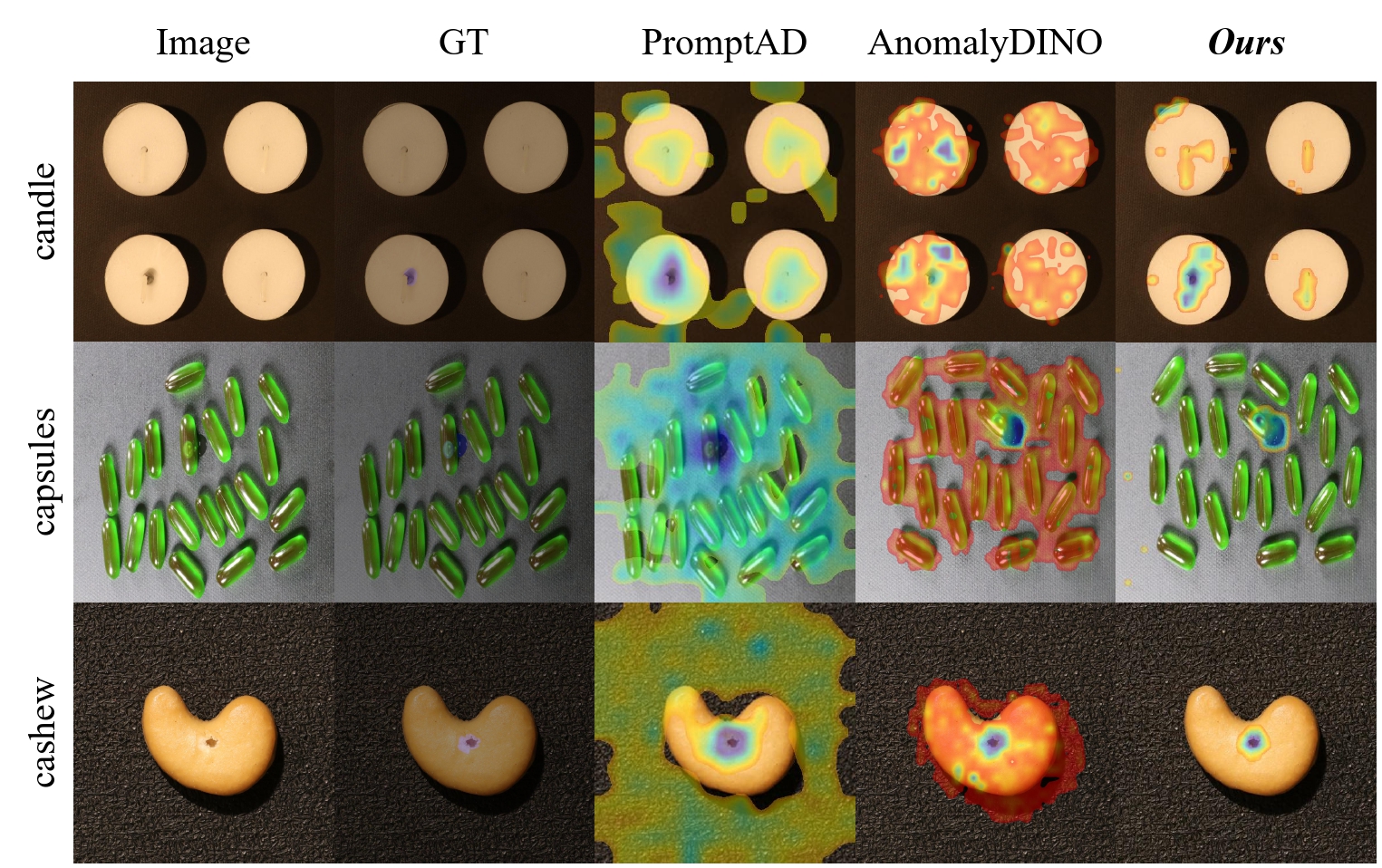}
    \vspace{2pt}
    \textbf{(a) VisA}
\end{minipage}
\hfill
\begin{minipage}[t]{0.49\textwidth}
    \centering
    \includegraphics[width=\linewidth]{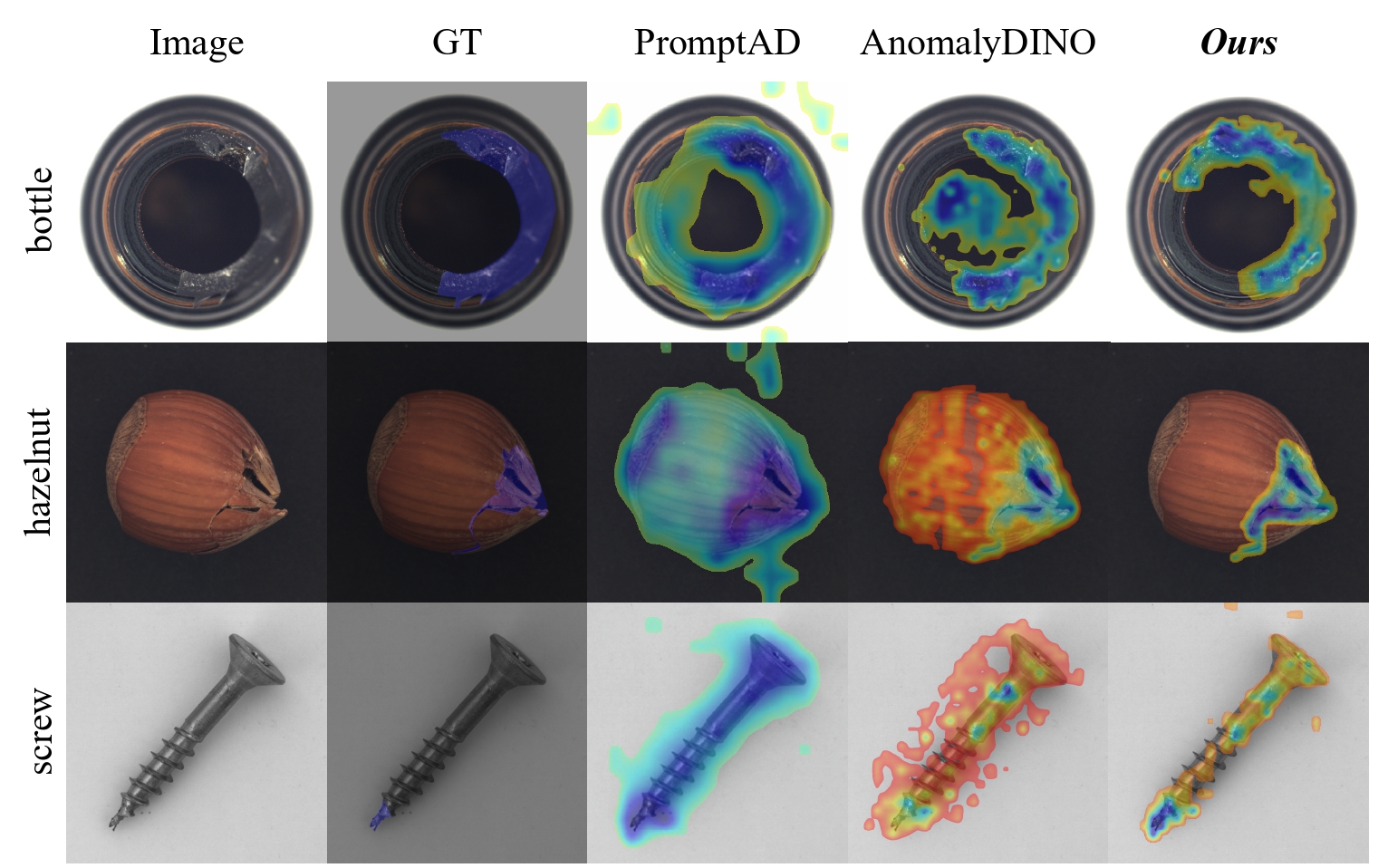}
    \vspace{2pt}
    \textbf{(b) MVTec-AD}
\end{minipage}
\vspace{3pt}
\caption{Qualitative comparison on VisA and MVTec-AD (1-shot). SubspaceAD produces sharper and more precise anomaly maps than PromptAD~\cite{li2024promptad} and AnomalyDINO~\cite{damm2025anomalydino}, with fewer false activations and better alignment with ground-truth defects across both datasets. More qualitative examples are provided in Appendix~\ref{sec:qualitative_results}.}
\label{fig:qualitative_combined}
\end{figure*}

Moreover, qualitative results in Fig.~\ref{fig:qualitative_combined} demonstrate that our method produces cleaner, sharper, and more spatially precise anomaly maps across both benchmarks. Further per-category results are included in Appendix~\ref{sec:per_category}, and representative failure modes are discussed in Appendix~\ref{sec:failure_cases}.

\subsection{Batched 0-Shot Performance}
\label{subsec:batched_0shot}
SubspaceAD is further evaluated under the \textit{batched 0-shot} setting, which differs fundamentally from prompt-based, zero-shot, or few-shot paradigms. Following the protocol of AnomalyDINO~\cite{damm2025anomalydino} and MuSc~\cite{li2024musc}, the entire test set for a category is used to construct the model, assuming that the majority of image patches are anomaly-free. Unlike memory–bank approaches that store all patches across images, SubspaceAD fits a single PCA subspace on all patch tokens extracted from the unlabeled test set of that category and computes anomaly scores based on reconstruction residuals.

Table~\ref{tab:batched_0shot_comparison} compares the performance of SubspaceAD against other batched 0-shot methods. SubspaceAD achieves state-of-the-art performance on the VisA dataset with an image-level AUROC of 94.1\%, matching the performance of MuSc and outperforming AnomalyDINO. On MVTec-AD, our method achieves a competitive 96.6\% AUROC.

\begin{table}[t]
\centering
\caption{Batched 0-shot anomaly detection. 
All values are image-level AUROC (\%). 
Best results are in \textbf{bold}, second-best are \underline{underlined}.}
\label{tab:batched_0shot_comparison}
\small
\setlength{\tabcolsep}{7pt}
\begin{tabular}{lcc}
\toprule
\textbf{Method} & \textbf{MVTec-AD} & \textbf{VisA} \\
\midrule
\multicolumn{3}{l}{\textit{Zero-shot methods}} \\
\quad WinCLIP~\cite{jeong2023winclip}              & 91.8 & 78.1 \\
\quad AnomalyCLIP~\cite{zhou2023anomalyclip}       & 91.5 & 82.1 \\
\midrule
\multicolumn{3}{l}{\textit{Batched zero-shot methods}} \\
\quad ACR~\cite{li2023zero}                        & 85.8 & -- \\
\quad MuSc~\cite{li2024musc}                       & \textbf{97.8} & \textbf{94.1} \\
\quad AnomalyDINO-S (448)~\cite{damm2025anomalydino} & 93.0 & 89.7 \\
\quad AnomalyDINO-S (672)~\cite{damm2025anomalydino} & 94.2 & 90.7 \\
\midrule
\textbf{SubspaceAD (ours)}                         & \underline{96.6} & \textbf{94.1} \\
\bottomrule
\end{tabular}
\end{table}

SubspaceAD differs from prior batched 0-shot approaches in how it models the unlabeled test set. AnomalyDINO builds a memory bank from all test patches, allowing anomalous regions to retrieve other anomalies as nearest neighbors and suppressing their anomaly scores. MuSc~\cite{li2024musc} alleviates this through mutual similarity filtering, but it requires dense cross-image comparisons and remains sensitive to contaminated categories. In contrast, SubspaceAD fits a single PCA model to all test tokens, where the principal components capture the shared, high-variance structure of normal data, while rare and uncorrelated anomalies reconstruct poorly and receive high anomaly scores. This compact, distribution-level modeling yields competitive performance on MVTec-AD (96.6\%) and achieves state-of-the-art results on VisA (94.1\%), showing that even in the batched 0-shot regime, a simple subspace is sufficient for strong anomaly discrimination.

\begin{figure}[hb]
\centering
\includegraphics[width=0.97\columnwidth]{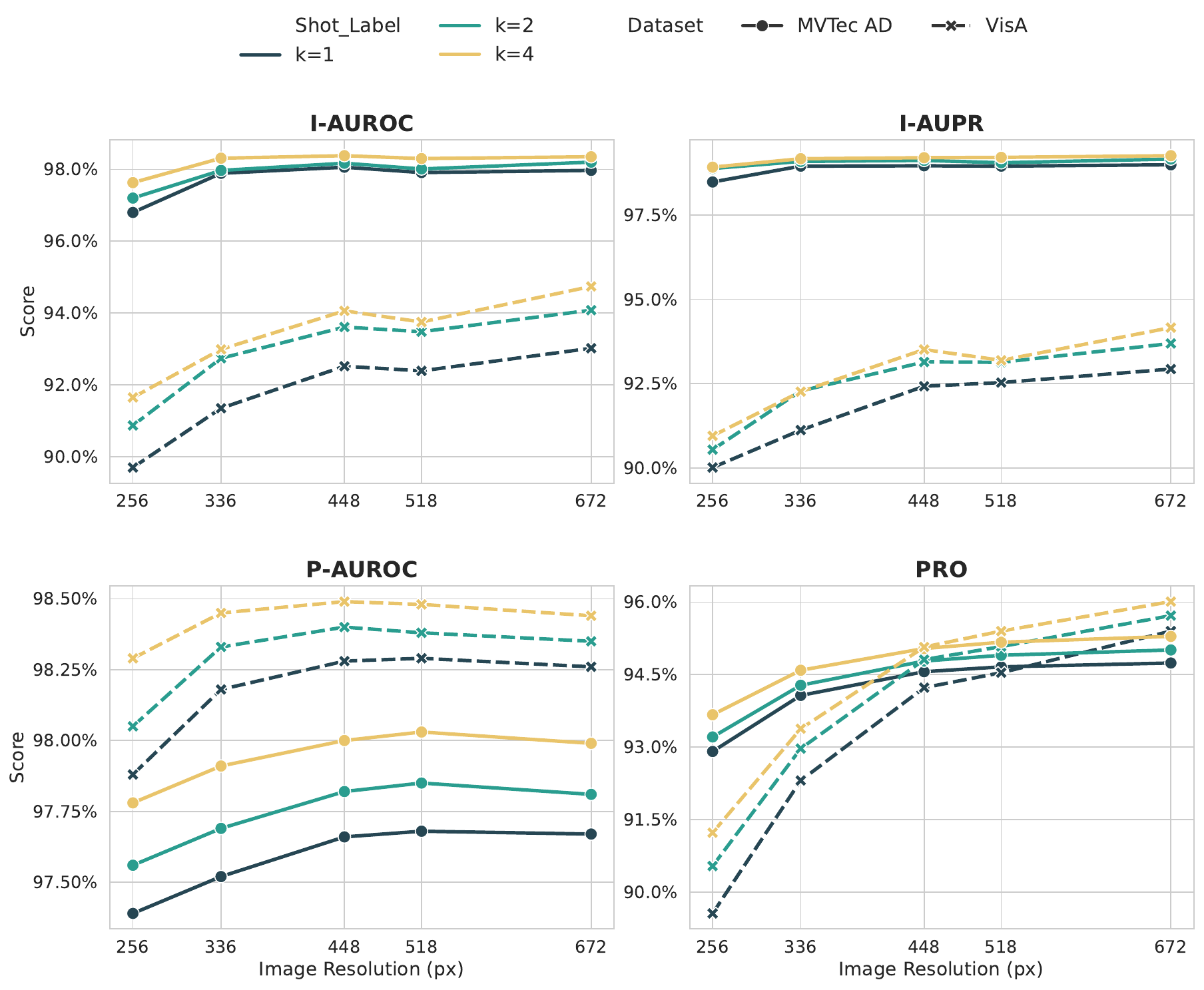}
\caption{Effect of image resolution on performance across both datasets. Performance peaks at 672~px on both MVTec-AD (solid) and VisA (dashed).}
\label{fig:ablation_resolution}
\end{figure}
\subsection{Ablation Study}
\label{subsec:ablations}
We analyze the impact of design choices in SubspaceAD, including (1) input image resolution, (2) layer aggregation strategy, (3) DINOv2 backbone scale, and (4) PCA explained-variance threshold~$\tau$. All experiments are performed on the MVTec-AD and VisA datasets.

\paragraph{Image Resolution}
Fig.~\ref{fig:ablation_resolution} shows the effect of input resolution. On VisA (dashed lines), performance improves from 256~px but quickly saturates, remaining largely stable between 448~px and 672~px. A similar trend is observed on MVTec-AD (solid lines), where all resolutions above 448~px perform comparably across metrics. These results indicate that the method is largely robust to resolution once sufficient spatial detail is available.

\paragraph{Layer Aggregation}
Table~\ref{tab:ablation_layers} compares aggregation strategies considering 4-shot results. Using only the last layer leads to a substantial performance drop (89.3\% I-AUROC on VisA). Averaging the final 7 layers (34-40) achieves better results but remains suboptimal. The \textit{Mean-pool (Middle-7)} configuration, averaging layers 22-28, delivers the best overall performance, reaching 95.0\% PRO on MVTec-AD and 94.1\% I-AUROC on VisA. While \textit{Concat (Middle-7)} yields a higher I-AUROC on MVTec-AD (98.6\%), our chosen pooling method provides the most robust and discriminative representation across both benchmarks.

\begin{table}[ht]
\centering
\caption{Evaluation of feature aggregation strategies (4-shot). The selected approach \textit{Mean-pool (Middle-7)} achieves the best overall performance, with image-level AUROC and PRO given in (\%).}
\label{tab:ablation_layers}
\small
\setlength{\tabcolsep}{5pt}
\begin{tabular}{l@{\hspace{1em}}cc@{\hspace{1em}}cc}
\toprule
& \multicolumn{2}{c}{\textbf{MVTec-AD}} & \multicolumn{2}{c}{\textbf{VisA}} \\
\cmidrule(lr){2-3} \cmidrule(lr){4-5}
\textbf{Strategy} & I-AUROC & PRO & I-AUROC & PRO \\
\midrule
\textbf{Mean-pool (Middle-7)} & 98.4 & \textbf{95.0} & \textbf{94.1} & \textbf{95.1} \\
Mean-pool (Final-7) & 98.2 & 93.3 & 91.9 & 93.5 \\
Concat (Middle-7) & \textbf{98.6} &\textbf{ 95.0} & 93.8 & 95.0 \\
Last layer only & 97.6 & 92.5 & 89.3 & 91.2 \\
\bottomrule
\end{tabular}
\end{table}

\begin{figure}[ht]
\centering
\includegraphics[width=0.97\columnwidth]{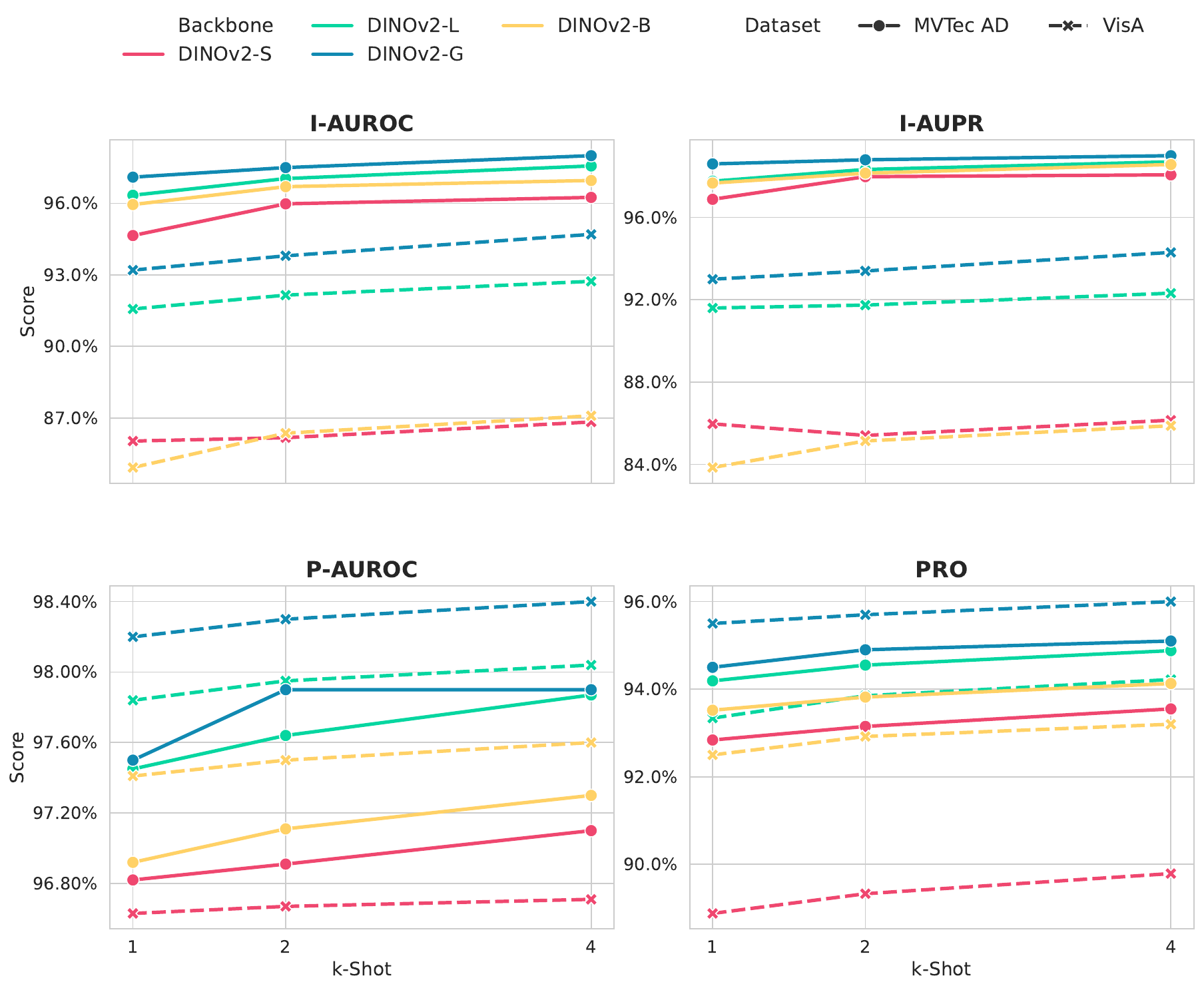}
\caption{Impact of backbone scale on SubspaceAD performance on both datasets. Performance improves with increasing model capacity, indicating that richer foundation features directly enhance few-shot anomaly detection.}
\label{fig:ablation_backbone}
\end{figure}

\vspace{3pt}
\paragraph{Backbone Scale} SubspaceAD is evaluated using four DINOv2 backbones: ViT-S/14, ViT-B/14, ViT-L/14, and ViT-G/14, as illustrated in Fig.~\ref{fig:ablation_backbone}. While larger backbones consistently yield better performance, practical deployment constraints (e.g., edge-device memory and latency) are governed primarily by the chosen encoder. Because SubspaceAD adds negligible computational overhead, it can be readily deployed with smaller, edge-friendly variants like DINOv2-S/14. For completeness, we also report results with DINOv3 backbones (Appendix~\ref{sec:dinov3-comp}), performing slightly worse in this setting.
\\

\paragraph{Explained Variance $\tau$}
Table~\ref{tab:ablation_variance} shows the effect of the PCA explained-variance threshold $\tau$. Performance remains stable for $\tau \in [0.95,0.99]$, showing that SubspaceAD is not overly sensitive to this hyperparameter. However, the $\tau=0.99$ threshold yields the best results for VisA and the strongest localization performance (PRO) on MVTec-AD. Consequently, we adopt $\tau=0.99$ for all experiments. As expected, setting $\tau=1.00$ (using the full feature space) causes performance to drop sharply, confirming that anomalies primarily reside in the residual subspace.

\begin{table}[ht]
\centering
\caption{Analysis of PCA explained variance $\tau$. Image-level AUROC (I-AUROC) and PRO (\%) are given for $k\!\in\!\{1,2,4\}$ on both MVTec-AD and VisA datasets. Best results are in \textbf{bold}.}
\label{tab:ablation_variance}
\small
\setlength{\tabcolsep}{3pt}
\begin{tabular}{lcccccc}
\toprule
& \multicolumn{3}{c}{\textbf{MVTec-AD}} & \multicolumn{3}{c}{\textbf{VisA}} \\
\cmidrule(lr){2-4}\cmidrule(lr){5-7}
$\tau$ & $k{=}1$ & $k{=}2$ & $k{=}4$ & $k{=}1$ & $k{=}2$ & $k{=}4$ \\
\midrule
\multicolumn{7}{l}{\textit{I-AUROC (\%)}}\\
0.95 & 98.0 & \textbf{98.2} & 98.4 & 92.2 & 93.4 & 93.7 \\
0.96 & \textbf{98.1} & \textbf{98.2} & 98.4 & 92.4 & 93.5 & 93.9 \\
0.97 & \textbf{98.1} & \textbf{98.2} & \textbf{98.5} & 92.4 & \textbf{93.6} & 93.9 \\
0.99 & \textbf{98.1} & \textbf{98.2} & 98.4 & \textbf{92.5} & \textbf{93.6} & \textbf{94.1} \\
1.00 & 45.6 & 41.4 & 40.5 & 35.3 & 37.9 & 38.3 \\
\midrule
\multicolumn{7}{l}{\textit{PRO (\%)}}\\
0.95 & 94.4 & 94.7 & 94.9 & 93.9 & 94.3 & 94.6 \\
0.96 & 94.4 & 94.7 & \textbf{95.0} & 94.0 & 94.4 & 94.8 \\
0.97 & 94.5 & 94.7 & \textbf{95.0} & 94.1 & 94.5 & 94.8 \\
0.99 & \textbf{94.6} & \textbf{94.8} & \textbf{95.0} & \textbf{94.2} & \textbf{94.8} & \textbf{95.1} \\
1.00 & 55.2 & 52.7 & 52.6 & 48.3 & 48.6 & 49.2 \\
\bottomrule
\end{tabular}
\end{table}
\paragraph{Summary}
The ablations show that performance is highly dependent on specific design choices. Model scale and the feature aggregation strategy have the strongest influence, followed closely by input resolution. The PCA threshold, in contrast, acts primarily as a fine-tuner. Overall, strong foundation features combined with a simple statistical subspace are sufficient for high-performing few-shot anomaly detection.
\section{Conclusion}
\label{sec:conclusion}
This paper introduced SubspaceAD, a training-free framework for few-shot visual anomaly detection that leverages the representational strength of vision foundation models. By extracting patch-level features from a frozen DINOv2-G encoder and modeling normal variation through a simple PCA subspace, the method detects anomalies via reconstruction residuals without requiring memory banks, auxiliary datasets, prompt tuning, or any form of training. Despite its simple formulation, SubspaceAD achieves state-of-the-art performance across one- and few-shot settings on both MVTec-AD and VisA datasets, demonstrating that complex architectures and multi-stage optimization are unnecessary when expressive feature representations are available.

\section*{Acknowledgements}
\raggedright
\small{
This work is supported by the ADVISOR ITEA 241007 project.
}
{
    \small
    \bibliographystyle{ieeenat_fullname}
    \bibliography{main}
}

\clearpage
\setcounter{page}{1}
\maketitlesupplementary
\appendix

\section{Per-Category Few-Shot Results}
\label{sec:per_category}
Tables~\ref{tab:mvtec_detailed_full} and~\ref{tab:visa_detailed_full} provide per-category few-shot performance on the MVTec-AD and VisA datasets.

\paragraph{MVTec-AD (Table~\ref{tab:mvtec_detailed_full})}
\textit{SubspaceAD} achieves consistently strong performance across nearly all MVTec-AD categories. Even with only one normal image, several categories (e.g., Bottle, Carpet, Grid, Leather, Tile, Toothbrush) reach perfect or near-perfect image-level AUROC, indicating robust capture of normal appearance from one example.
The \textit{Transistor} category is a notable exception, with a 1-shot Pixel PRO score of 68.9\%. This limitation is due to the characteristic of patch-based visual anomaly detection, which focuses on local appearance and therefore does not explicitly account for logical or structural anomalies, such as missing or misplaced components.

\paragraph{VisA (Table~\ref{tab:visa_detailed_full})}
\textit{SubspaceAD} exhibits greater performance variability across VisA categories, which is expected given the dataset's higher visual diversity and complex backgrounds~\cite{zou2022spotvisa}. Categories such as \textit{Cashew} and \textit{Chewing Gum} achieve excellent results, with 1-shot image-level AUROC scores of 97.7\% and 99.2\%, respectively.
Conversely, categories like \textit{Macaroni2} (80.4\%) and \textit{PCB3} (86.5\%) are more challenging. This degradation primarily stems from background artifacts that resemble true defects and from the high intra-class variability of normal samples. Both factors hinder patch-based methods from forming a compact subspace of normality from only a few shots.

\begin{table*}[ht]
\caption{Detailed few-shot anomaly segmentation results of \textit{SubspaceAD} on the MVTec AD dataset. We report mean Image AUROC (\%), Image AUPR (\%), Pixel AUROC (\%), and Pixel PRO (\%) results.}
\label{tab:mvtec_detailed_full}
\centering
\resizebox{\textwidth}{!}{
\begin{tabular}{@{}l|rrrr|rrrr|rrrr@{}}
\toprule
\multirow{2}{*}{Category} & \multicolumn{4}{c|}{1-shot} & \multicolumn{4}{c|}{2-shot} & \multicolumn{4}{c@{}}{4-shot} \\ \cmidrule(l){2-13} 
 & Img AUROC & Img AUPR & Pxl AUROC & PRO & Img AUROC & Img AUPR & Pxl AUROC & PRO & Img AUROC & Img AUPR & Pxl AUROC & PRO \\ \midrule
Bottle & 100.0 $\pm$ 0.0 & 100.0 $\pm$ 0.0 & 98.6 $\pm$ 0.0 & 96.5 $\pm$ 0.1 & 100.0 $\pm$ 0.0 & 100.0 $\pm$ 0.0 & 98.7 $\pm$ 0.0 & 96.7 $\pm$ 0.1 & 100.0 $\pm$ 0.0 & 100.0 $\pm$ 0.0 & 98.7 $\pm$ 0.0 & 96.7 $\pm$ 0.1 \\
Cable & 92.1 $\pm$ 1.4 & 95.5 $\pm$ 0.7 & 95.6 $\pm$ 0.2 & 90.3 $\pm$ 0.7 & 92.7 $\pm$ 1.2 & 96.2 $\pm$ 0.6 & 95.6 $\pm$ 0.2 & 90.6 $\pm$ 0.5 & 93.5 $\pm$ 0.7 & 96.7 $\pm$ 0.5 & 95.7 $\pm$ 0.2 & 90.9 $\pm$ 0.4 \\
Capsule & 88.1 $\pm$ 11.3 & 96.8 $\pm$ 3.9 & 98.0 $\pm$ 0.2 & 96.7 $\pm$ 0.5 & 92.7 $\pm$ 3.6 & 98.4 $\pm$ 0.9 & 98.2 $\pm$ 0.1 & 97.1 $\pm$ 0.2 & 93.4 $\pm$ 4.8 & 98.5 $\pm$ 1.2 & 98.3 $\pm$ 0.1 & 97.3 $\pm$ 0.2 \\
Carpet & 100.0 $\pm$ 0.0 & 100.0 $\pm$ 0.0 & 99.2 $\pm$ 0.0 & 98.2 $\pm$ 0.1 & 100.0 $\pm$ 0.0 & 100.0 $\pm$ 0.0 & 99.2 $\pm$ 0.0 & 98.2 $\pm$ 0.0 & 100.0 $\pm$ 0.0 & 100.0 $\pm$ 0.0 & 99.3 $\pm$ 0.0 & 98.3 $\pm$ 0.0 \\
Grid & 100.0 $\pm$ 0.0 & 100.0 $\pm$ 0.0 & 99.5 $\pm$ 0.0 & 98.1 $\pm$ 0.1 & 100.0 $\pm$ 0.0 & 100.0 $\pm$ 0.0 & 99.5 $\pm$ 0.0 & 98.2 $\pm$ 0.1 & 100.0 $\pm$ 0.0 & 100.0 $\pm$ 0.0 & 99.5 $\pm$ 0.0 & 98.3 $\pm$ 0.1 \\
Hazelnut & 97.6 $\pm$ 3.4 & 98.6 $\pm$ 2.0 & 99.5 $\pm$ 0.2 & 97.1 $\pm$ 1.2 & 97.3 $\pm$ 6.0 & 98.1 $\pm$ 4.3 & 99.6 $\pm$ 0.1 & 97.8 $\pm$ 0.6 & 99.4 $\pm$ 1.4 & 99.6 $\pm$ 0.8 & 99.6 $\pm$ 0.0 & 98.1 $\pm$ 0.3 \\
Leather & 100.0 $\pm$ 0.0 & 100.0 $\pm$ 0.0 & 98.9 $\pm$ 0.0 & 98.3 $\pm$ 0.1 & 100.0 $\pm$ 0.0 & 100.0 $\pm$ 0.0 & 98.9 $\pm$ 0.0 & 98.3 $\pm$ 0.1 & 100.0 $\pm$ 0.0 & 100.0 $\pm$ 0.0 & 98.9 $\pm$ 0.0 & 98.3 $\pm$ 0.1 \\
Metal Nut & 100.0 $\pm$ 0.0 & 100.0 $\pm$ 0.0 & 97.2 $\pm$ 0.9 & 95.9 $\pm$ 0.9 & 100.0 $\pm$ 0.0 & 100.0 $\pm$ 0.0 & 97.8 $\pm$ 0.2 & 96.5 $\pm$ 0.2 & 100.0 $\pm$ 0.0 & 100.0 $\pm$ 0.0 & 97.8 $\pm$ 0.1 & 96.5 $\pm$ 0.2 \\
Pill & 96.2 $\pm$ 0.8 & 99.2 $\pm$ 0.2 & 95.5 $\pm$ 0.5 & 97.8 $\pm$ 0.1 & 96.6 $\pm$ 0.5 & 99.4 $\pm$ 0.1 & 95.8 $\pm$ 0.2 & 97.9 $\pm$ 0.1 & 97.2 $\pm$ 0.8 & 99.4 $\pm$ 0.2 & 96.2 $\pm$ 0.4 & 98.1 $\pm$ 0.2 \\
Screw & 88.4 $\pm$ 1.7 & 95.4 $\pm$ 0.9 & 99.1 $\pm$ 0.1 & 96.2 $\pm$ 0.3 & 90.2 $\pm$ 2.8 & 96.2 $\pm$ 1.2 & 99.2 $\pm$ 0.1 & 96.7 $\pm$ 0.3 & 91.9 $\pm$ 1.1 & 96.9 $\pm$ 0.5 & 99.3 $\pm$ 0.0 & 97.0 $\pm$ 0.1 \\
Tile & 100.0 $\pm$ 0.1 & 100.0 $\pm$ 0.0 & 97.7 $\pm$ 0.1 & 94.8 $\pm$ 0.2 & 100.0 $\pm$ 0.0 & 100.0 $\pm$ 0.0 & 97.7 $\pm$ 0.1 & 94.7 $\pm$ 0.2 & 100.0 $\pm$ 0.1 & 100.0 $\pm$ 0.0 & 97.7 $\pm$ 0.1 & 94.6 $\pm$ 0.1 \\
Toothbrush & 98.8 $\pm$ 1.4 & 99.5 $\pm$ 0.6 & 98.9 $\pm$ 0.5 & 97.0 $\pm$ 0.6 & 96.7 $\pm$ 3.4 & 98.7 $\pm$ 1.3 & 98.8 $\pm$ 0.5 & 96.9 $\pm$ 0.7 & 98.8 $\pm$ 0.7 & 99.6 $\pm$ 0.3 & 99.2 $\pm$ 0.2 & 97.5 $\pm$ 0.4 \\
Transistor & 96.6 $\pm$ 0.8 & 94.5 $\pm$ 1.4 & 90.4 $\pm$ 1.1 & 68.9 $\pm$ 1.4 & 96.7 $\pm$ 1.3 & 94.7 $\pm$ 2.3 & 92.1 $\pm$ 0.8 & 71.5 $\pm$ 1.2 & 96.3 $\pm$ 2.4 & 94.7 $\pm$ 3.0 & 92.6 $\pm$ 1.1 & 72.7 $\pm$ 1.5 \\
Wood & 99.6 $\pm$ 0.2 & 99.9 $\pm$ 0.0 & 96.7 $\pm$ 0.3 & 96.5 $\pm$ 0.2 & 99.7 $\pm$ 0.1 & 99.9 $\pm$ 0.0 & 96.8 $\pm$ 0.5 & 96.7 $\pm$ 0.2 & 99.7 $\pm$ 0.1 & 99.9 $\pm$ 0.0 & 96.8 $\pm$ 0.2 & 96.6 $\pm$ 0.1 \\
Zipper & 99.1 $\pm$ 0.5 & 99.8 $\pm$ 0.1 & 98.3 $\pm$ 0.2 & 95.4 $\pm$ 0.6 & 99.5 $\pm$ 0.2 & 99.9 $\pm$ 0.0 & 98.4 $\pm$ 0.2 & 95.8 $\pm$ 0.4 & 99.5 $\pm$ 0.2 & 99.9 $\pm$ 0.1 & 98.6 $\pm$ 0.1 & 96.2 $\pm$ 0.3 \\
\midrule
\textbf{Mean} & \textbf{97.1 $\pm$ 0.9} & \textbf{98.6 $\pm$ 0.4} & \textbf{97.5 $\pm$ 0.1} & \textbf{94.5 $\pm$ 0.2} & \textbf{97.5 $\pm$ 0.7} & \textbf{98.8 $\pm$ 0.4} & \textbf{97.8 $\pm$ 0.1} & \textbf{94.9 $\pm$ 0.1} & \textbf{98.0 $\pm$ 0.4} & \textbf{99.0 $\pm$ 0.2} & \textbf{97.9 $\pm$ 0.1} & \textbf{95.1 $\pm$ 0.1} \\
\bottomrule
\end{tabular}%
}
\end{table*}

\begin{table*}[ht]
\caption{Detailed few-shot anomaly segmentation results of \textit{SubspaceAD} on the VisA dataset. We report mean Image AUROC (\%), Image AUPR (\%), Pixel AUROC (\%), and Pixel PRO (\%) results.}
\label{tab:visa_detailed_full}
\centering
\resizebox{\textwidth}{!}{
\begin{tabular}{@{}l|rrrr|rrrr|rrrr@{}}
\toprule
\multirow{2}{*}{Category} & \multicolumn{4}{c|}{1-shot} & \multicolumn{4}{c|}{2-shot} & \multicolumn{4}{c@{}}{4-shot} \\ \cmidrule(l){2-13} 
 & Img AUROC & Img AUPR & Pxl AUROC & PRO & Img AUROC & Img AUPR & Pxl AUROC & PRO & Img AUROC & Img AUPR & Pxl AUROC & PRO \\ \midrule
Candle & 94.4 $\pm$ 0.8 & 94.2 $\pm$ 0.7 & 99.4 $\pm$ 0.0 & 97.9 $\pm$ 0.2 & 94.2 $\pm$ 0.5 & 93.8 $\pm$ 0.7 & 99.4 $\pm$ 0.0 & 98.1 $\pm$ 0.1 & 93.9 $\pm$ 0.9 & 93.4 $\pm$ 1.1 & 99.4 $\pm$ 0.0 & 98.1 $\pm$ 0.1 \\
Capsules & 96.5 $\pm$ 0.7 & 97.9 $\pm$ 0.4 & 98.5 $\pm$ 0.2 & 97.0 $\pm$ 0.3 & 96.8 $\pm$ 0.3 & 98.1 $\pm$ 0.2 & 98.6 $\pm$ 0.1 & 97.2 $\pm$ 0.3 & 97.0 $\pm$ 0.9 & 98.2 $\pm$ 0.6 & 98.6 $\pm$ 0.1 & 97.3 $\pm$ 0.3 \\
Cashew & 97.7 $\pm$ 0.7 & 98.9 $\pm$ 0.3 & 99.0 $\pm$ 0.0 & 98.5 $\pm$ 0.2 & 97.9 $\pm$ 1.3 & 99.0 $\pm$ 0.6 & 99.2 $\pm$ 0.1 & 98.6 $\pm$ 0.1 & 98.7 $\pm$ 0.8 & 99.4 $\pm$ 0.4 & 99.3 $\pm$ 0.1 & 98.7 $\pm$ 0.1 \\
Chewing Gum & 99.2 $\pm$ 0.1 & 99.6 $\pm$ 0.1 & 99.6 $\pm$ 0.0 & 96.0 $\pm$ 0.2 & 99.1 $\pm$ 0.1 & 99.6 $\pm$ 0.0 & 99.6 $\pm$ 0.0 & 96.3 $\pm$ 0.1 & 99.1 $\pm$ 0.0 & 99.5 $\pm$ 0.0 & 99.6 $\pm$ 0.0 & 96.2 $\pm$ 0.1 \\
Fryum & 97.0 $\pm$ 0.6 & 98.7 $\pm$ 0.3 & 96.5 $\pm$ 0.3 & 95.4 $\pm$ 0.5 & 97.9 $\pm$ 0.1 & 99.1 $\pm$ 0.1 & 96.8 $\pm$ 0.1 & 95.5 $\pm$ 0.3 & 98.1 $\pm$ 0.3 & 99.2 $\pm$ 0.1 & 97.0 $\pm$ 0.1 & 95.7 $\pm$ 0.3 \\
Macaroni1 & 92.0 $\pm$ 1.1 & 91.5 $\pm$ 1.4 & 99.6 $\pm$ 0.0 & 95.8 $\pm$ 0.3 & 92.6 $\pm$ 1.4 & 92.1 $\pm$ 1.4 & 99.7 $\pm$ 0.0 & 96.2 $\pm$ 0.4 & 94.2 $\pm$ 1.6 & 93.7 $\pm$ 1.3 & 99.8 $\pm$ 0.0 & 96.8 $\pm$ 0.4 \\
Macaroni2 & 80.4 $\pm$ 5.1 & 76.2 $\pm$ 8.0 & 99.7 $\pm$ 0.0 & 97.5 $\pm$ 0.5 & 81.4 $\pm$ 5.4 & 77.1 $\pm$ 9.0 & 99.7 $\pm$ 0.1 & 97.8 $\pm$ 0.2 & 84.7 $\pm$ 3.9 & 81.8 $\pm$ 6.5 & 99.8 $\pm$ 0.0 & 98.0 $\pm$ 0.2 \\
PCB1 & 92.4 $\pm$ 3.2 & 90.8 $\pm$ 2.6 & 99.3 $\pm$ 0.0 & 95.0 $\pm$ 0.5 & 92.8 $\pm$ 1.4 & 91.2 $\pm$ 1.0 & 99.3 $\pm$ 0.0 & 95.1 $\pm$ 0.3 & 93.2 $\pm$ 0.9 & 91.6 $\pm$ 0.9 & 99.4 $\pm$ 0.0 & 95.3 $\pm$ 0.1 \\
PCB2 & 88.7 $\pm$ 2.2 & 86.7 $\pm$ 2.1 & 96.9 $\pm$ 0.3 & 91.4 $\pm$ 0.6 & 90.5 $\pm$ 1.0 & 87.5 $\pm$ 1.6 & 97.1 $\pm$ 0.1 & 92.2 $\pm$ 0.3 & 91.7 $\pm$ 1.3 & 88.9 $\pm$ 1.7 & 97.2 $\pm$ 0.0 & 92.7 $\pm$ 0.2 \\
PCB3 & 86.5 $\pm$ 1.8 & 86.1 $\pm$ 2.2 & 95.0 $\pm$ 0.4 & 90.4 $\pm$ 0.8 & 88.0 $\pm$ 2.6 & 87.4 $\pm$ 3.1 & 95.2 $\pm$ 0.1 & 91.1 $\pm$ 0.6 & 89.9 $\pm$ 2.3 & 89.8 $\pm$ 2.0 & 95.4 $\pm$ 0.2 & 91.8 $\pm$ 0.4 \\
PCB4 & 98.3 $\pm$ 0.7 & 98.1 $\pm$ 0.8 & 96.4 $\pm$ 0.4 & 92.2 $\pm$ 1.1 & 98.1 $\pm$ 0.6 & 97.8 $\pm$ 0.8 & 96.6 $\pm$ 0.3 & 92.3 $\pm$ 0.9 & 98.4 $\pm$ 0.4 & 98.0 $\pm$ 0.6 & 96.8 $\pm$ 0.2 & 92.4 $\pm$ 1.0 \\
Pipe Fryum & 95.7 $\pm$ 2.3 & 97.6 $\pm$ 1.3 & 98.7 $\pm$ 0.1 & 98.3 $\pm$ 0.1 & 96.1 $\pm$ 1.9 & 97.8 $\pm$ 1.0 & 98.9 $\pm$ 0.1 & 98.4 $\pm$ 0.1 & 97.8 $\pm$ 0.3 & 98.7 $\pm$ 0.2 & 99.0 $\pm$ 0.0 & 98.5 $\pm$ 0.1 \\
\midrule
\textbf{Mean} & \textbf{93.2 $\pm$ 0.8} & \textbf{93.0 $\pm$ 0.8} & \textbf{98.2 $\pm$ 0.1} & \textbf{95.5 $\pm$ 0.1} & \textbf{93.8 $\pm$ 0.4} & \textbf{93.4 $\pm$ 0.7} & \textbf{98.3 $\pm$ 0.0} & \textbf{95.7 $\pm$ 0.1} & \textbf{94.7 $\pm$ 0.2} & \textbf{94.3 $\pm$ 0.5} & \textbf{98.4 $\pm$ 0.0} & \textbf{96.0 $\pm$ 0.1} \\
\bottomrule
\end{tabular}%
}
\end{table*}

\section{Performance of DINOv3 Backbones}
\label{sec:dinov3-comp}
For completeness, we report the few-shot results using the DINOv3-7B backbone~\cite{simeoni2025dinov3} in Table~\ref{tab:dinov3_few_shot}. These results are obtained using 4096-dimensional features, extracted from layers 22--28 with 16$\times$16 patch tokens. SubspaceAD$_{448}$ and SubspaceAD$_{672}$ denote models evaluated at input image resolutions of 448$\times$448 and 672$\times$672 pixels, respectively. As shown in Table~\ref{tab:dinov3_few_shot}, the DINOv3 backbone does not improve over the DINOv2-G backbone used in the main paper (see Table~\ref{tab:main_comparison}), with DINOv2-G remaining stronger overall, especially for localization metrics.

\begin{table*}[t]
\centering
\caption{Few-shot anomaly detection and localization results using the DINOv3-7B backbone. This backbone is consistently outperformed by DINOv2-G (see Table~\ref{tab:main_comparison} in the main manuscript).}
\label{tab:dinov3_few_shot}
\small
\begin{tabular}{@{}llcccccccc@{}}
\toprule
 &  & \multicolumn{4}{c}{\textbf{MVTec-AD}} & \multicolumn{4}{c}{\textbf{VisA}} \\
\cmidrule(lr){3-6} \cmidrule(lr){7-10}
 &  & \multicolumn{2}{c}{Image-level} & \multicolumn{2}{c}{Pixel-level} & \multicolumn{2}{c}{Image-level} & \multicolumn{2}{c}{Pixel-level} \\
\cmidrule(lr){3-4} \cmidrule(lr){5-6} \cmidrule(lr){7-8} \cmidrule(lr){9-10}
\textbf{Setup} & \textbf{Method} & AUROC & AUPR & AUROC & PRO & AUROC & AUPR & AUROC & PRO \\
\midrule

\multirow{2}{*}{1-shot}
 & SubspaceAD$_{448}$ (DINOv3) & 96.4 & 98.0 & 97.3 & 94.0 & 91.8 & 91.9 & 98.0 & 93.8 \\
 & SubspaceAD$_{672}$ (DINOv3) & 96.2 & 97.9 & 97.3 & 94.2 & 92.8 & 92.9 & 98.0 & 94.7 \\
\midrule

\multirow{2}{*}{2-shot}
 & SubspaceAD$_{448}$ (DINOv3) & 97.1 & 98.3 & 97.6 & 94.5 & 92.9 & 92.7 & 98.2 & 94.3 \\
 & SubspaceAD$_{672}$ (DINOv3) & 96.8 & 98.2 & 97.5 & 94.7 & 93.7 & 93.6 & 98.1 & 95.1 \\
\midrule

\multirow{2}{*}{4-shot}
 & SubspaceAD$_{448}$ (DINOv3) & 97.6 & 98.6 & 97.8 & 94.8 & 93.7 & 93.4 & 98.3 & 94.7 \\
 & SubspaceAD$_{672}$ (DINOv3) & 97.4 & 98.5 & 97.8 & 95.0 & 94.6 & 94.5 & 98.3 & 95.3 \\
\bottomrule
\end{tabular}
\end{table*}

\section{Full-Shot Setting Analysis}
\label{sec:full_shot}
A comparison in the full-shot setting, where all available normal training samples are used to build the model of normality, is provided in Table~\ref{tab:full_shot_comparison}. On MVTec-AD, AnomalyDINO obtains slightly higher image-level performance, with 99.5\% I-AUROC compared to 99.2\% for \textit{SubspaceAD}. Both methods achieve the same Pixel AUROC of 98.2\%, while \textit{SubspaceAD} obtains a higher Pixel PRO score of 95.6\% compared to 95.0\%.

On the more challenging VisA dataset, \textit{SubspaceAD} surpasses AnomalyDINO across all reported metrics, including image-level AUROC (98.2\% vs. 97.6\%), image-level AUPR (98.3\% vs. 98.0\%), pixel-level AUROC (99.1\% vs. 98.8\%), and Pixel PRO (96.9\% vs. 96.1\%). This is notable because AnomalyDINO stores dense patch-level features for every normal image and performs K-NN retrieval over these embeddings, typically on the order of $10^6$ feature vectors per category on VisA.
In contrast, \textit{SubspaceAD} performs a single forward pass and a lightweight subspace projection, requiring no feature memory banks.

\begin{table*}[t]
\centering
\caption{Comparison of anomaly detection and localization performance on MVTec-AD and VisA in the \textbf{full-shot} setting. Best results are in \textbf{bold}, and second-best are \underline{underlined}.}
\label{tab:full_shot_comparison}
\small
\begin{tabular}{@{}llcccccccc@{}}
\toprule
 &  & \multicolumn{4}{c}{\textbf{MVTec-AD}} & \multicolumn{4}{c}{\textbf{VisA}} \\
\cmidrule(lr){3-6} \cmidrule(lr){7-10}
 &  & \multicolumn{2}{c}{Image-level} & \multicolumn{2}{c}{Pixel-level} & \multicolumn{2}{c}{Image-level} & \multicolumn{2}{c}{Pixel-level} \\
\cmidrule(lr){3-4} \cmidrule(lr){5-6} \cmidrule(lr){7-8} \cmidrule(lr){9-10}
\textbf{Setup} & \textbf{Method} & AUROC & AUPR & AUROC & PRO & AUROC & AUPR & AUROC & PRO \\
\midrule
Full-shot & AnomalyDINO & \textbf{99.5} & \textbf{99.8} & \textbf{98.2} & \underline{95.0} & \underline{97.6} & \underline{98.0} & \underline{98.8} & \underline{96.1} \\
Full-shot & \textit{SubspaceAD} (Ours) & \underline{99.2} & \underline{99.6} & \textbf{98.2} & \textbf{95.6} & \textbf{98.2} & \textbf{98.3} & \textbf{99.1} & \textbf{96.9} \\
\bottomrule
\end{tabular}
\end{table*}

\section{Inference Time Analysis}
\label{sec:inference_time}
An analysis of inference speed is presented in Table~\ref{tab:runtime_comparison}. To ensure an algorithmic comparison independent of hardware, we further evaluate the standalone \textit{scoring head} latency (isolated from the backbone forward pass) in Table~\ref{tab:scoring_latency}.

\paragraph{End-to-End Latency}
Table~\ref{tab:runtime_comparison} reports the end-to-end inference times for various backbone configurations. Because these measurements were obtained on different hardware (NVIDIA A40 for AnomalyDINO vs. NVIDIA H100 for SubspaceAD), direct wall-clock speed comparisons are not appropriate. For both methods, the total inference time is primarily dominated by the forward pass of the frozen backbone. The key distinction is architectural: SubspaceAD achieves its performance using only the backbone and a lightweight projection, eliminating the need to store or search through feature memory banks.

\paragraph{Algorithmic Fairness (Scoring Head Only)}
To isolate the scoring mechanism from the backbone's overhead, we measured the time required to compute anomaly scores from extracted features on an H100. As shown in Table~\ref{tab:scoring_latency}, both methods exhibit comparable latency in the few-shot regime. SubspaceAD requires $\approx$74~ms per image, while AnomalyDINO requires $\approx$80~ms. While our subspace projection time is strictly invariant to the size of the support set once the PCA components are determined, the k-NN retrieval overhead of AnomalyDINO remains similarly marginal for the small values of $K$ evaluated here.

\begin{table}[ht]
\centering
\caption{Hardware-normalized scoring head latency (H100). We measure the time to process features \textit{after} extraction. SubspaceAD latency is invariant to $K$, whereas memory-bank retrieval scales with the number of shots.}
\label{tab:scoring_latency}
\small
\begin{tabular}{@{}l c S[table-format=2.1]@{}}
\toprule
\textbf{Method} & \textbf{Setting ($K$)} & {\textbf{Scoring Time (ms / img)}} \\
\midrule
AnomalyDINO & 1-shot & 80.1 \\
AnomalyDINO & 2-shot & 80.3 \\
AnomalyDINO & 4-shot & 80.5 \\
\midrule
\textbf{SubspaceAD (Ours)} & 1--4 shot & \textbf{74.1} \\
\bottomrule
\end{tabular}
\end{table}

\begin{table*}[t]
\centering
\caption{Inference time comparison with AnomalyDINO on MVTec-AD (1-shot, \textbf{448px}). 
Note that AnomalyDINO results are from the authors’ paper~\cite{damm2025anomalydino}, measured on an NVIDIA A40, 
while our measurements are obtained on an NVIDIA H100.}
\label{tab:runtime_comparison}
\small
\begin{tabular}{@{}l l S[table-format=4.1] l S[table-format=4.0]@{}}
\toprule
\textbf{Method} & \textbf{Backbone} & {\textbf{Params (M)}} & \textbf{GPU} & {\textbf{Time (ms / image)}} \\
\midrule
WinCLIP~\cite{jeong2023winclip} & CLIP ViT-B/16+ & 150.0 & NVIDIA T4 & 389\\
\midrule
AnomalyDINO~\cite{damm2025anomalydino} & DINOv2 ViT-S & 21.7 & NVIDIA A40 & 60 \\
AnomalyDINO~\cite{damm2025anomalydino} & DINOv2 ViT-B & 85.8 & NVIDIA A40 & 84 \\
AnomalyDINO~\cite{damm2025anomalydino} & DINOv2 ViT-L & 303.3 & NVIDIA A40 & 141 \\
\midrule
\multicolumn{5}{l}{\textit{SubspaceAD (Ours) -- DINOv2 Backbones (NVIDIA H100)}} \\
\textbf{SubspaceAD (Ours)} & DINOv2 ViT-S & 21.7 & NVIDIA H100 & 36 \\
\textbf{SubspaceAD (Ours)} & DINOv2 ViT-B & 85.8 & NVIDIA H100 & 56 \\
\textbf{SubspaceAD (Ours)} & DINOv2 ViT-L & 303.3 & NVIDIA H100 & 112 \\
\textbf{SubspaceAD (Ours)} & DINOv2 ViT-G & 1100.0 & NVIDIA H100 & 127 \\
\midrule
\multicolumn{5}{l}{\textit{SubspaceAD (Ours) -- DINOv3 Backbones (NVIDIA H100)}} \\
\textbf{SubspaceAD (Ours)} & DINOv3 ViT-S & 21.0 & NVIDIA H100 & 16 \\
\textbf{SubspaceAD (Ours)} & DINOv3 ViT-B & 86.0 & NVIDIA H100 & 21 \\
\textbf{SubspaceAD (Ours)} & DINOv3 ViT-7B & 7000.0 & NVIDIA H100 & 330 \\
\bottomrule
\end{tabular}
\end{table*}

\section{Additional Qualitative Results}
\label{sec:qualitative_results}
\begin{figure}[ht]
    \centering
    \includegraphics[width=\linewidth]{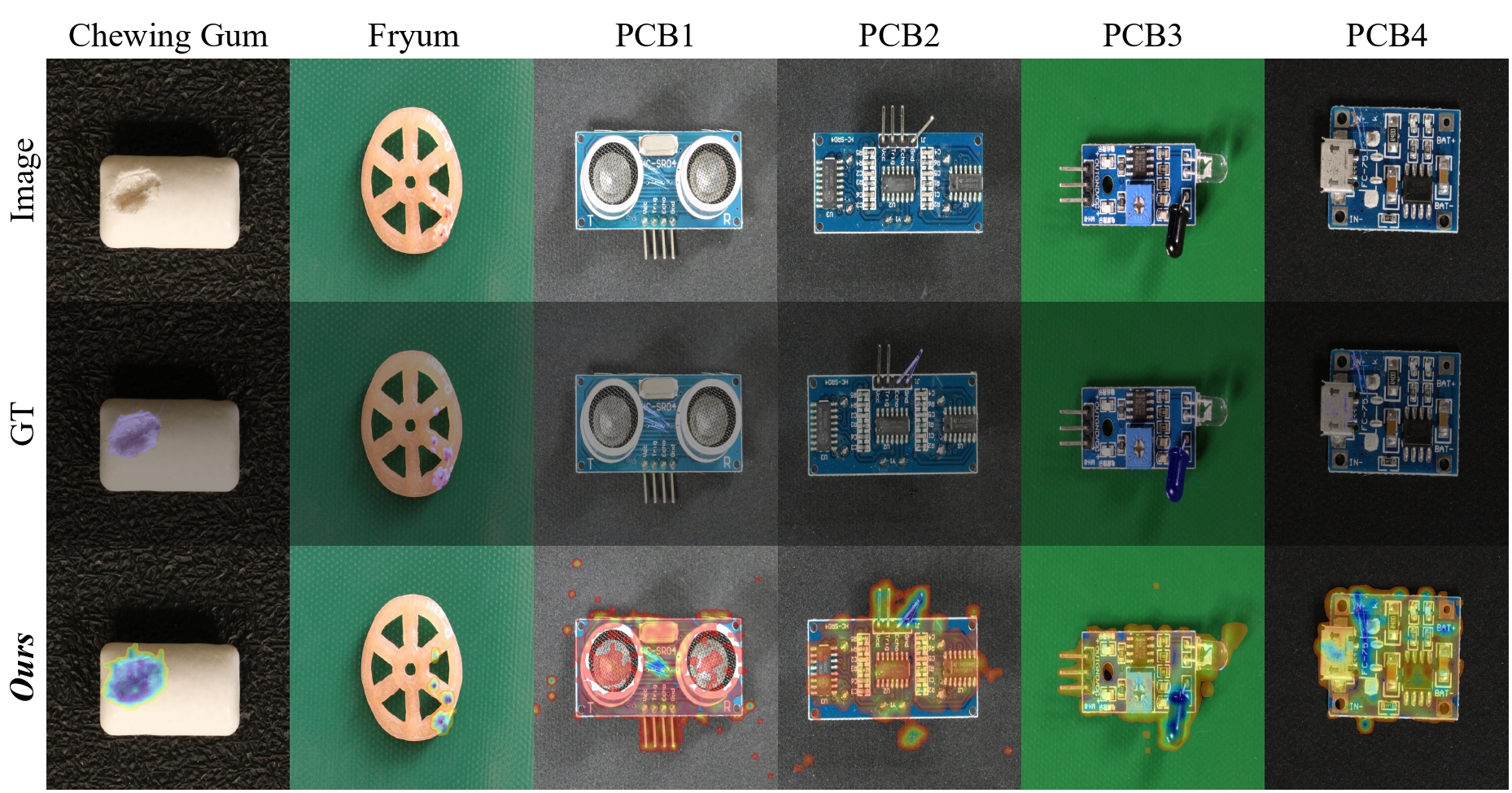}
    \caption{\textit{Additional qualitative results on VisA.}
    Examples from six categories (\textit{PCB1--4}, \textit{Chewing Gum}, \textit{Fryum}). 
    Rows show the input image, ground-truth mask, and our prediction. 
    \textit{SubspaceAD} accurately localizes both subtle texture anomalies and fine structural defects across diverse VisA domains.}
    \label{fig:visA_qualitative}
\end{figure}

\begin{figure}[ht]
    \centering
    \includegraphics[width=\linewidth]{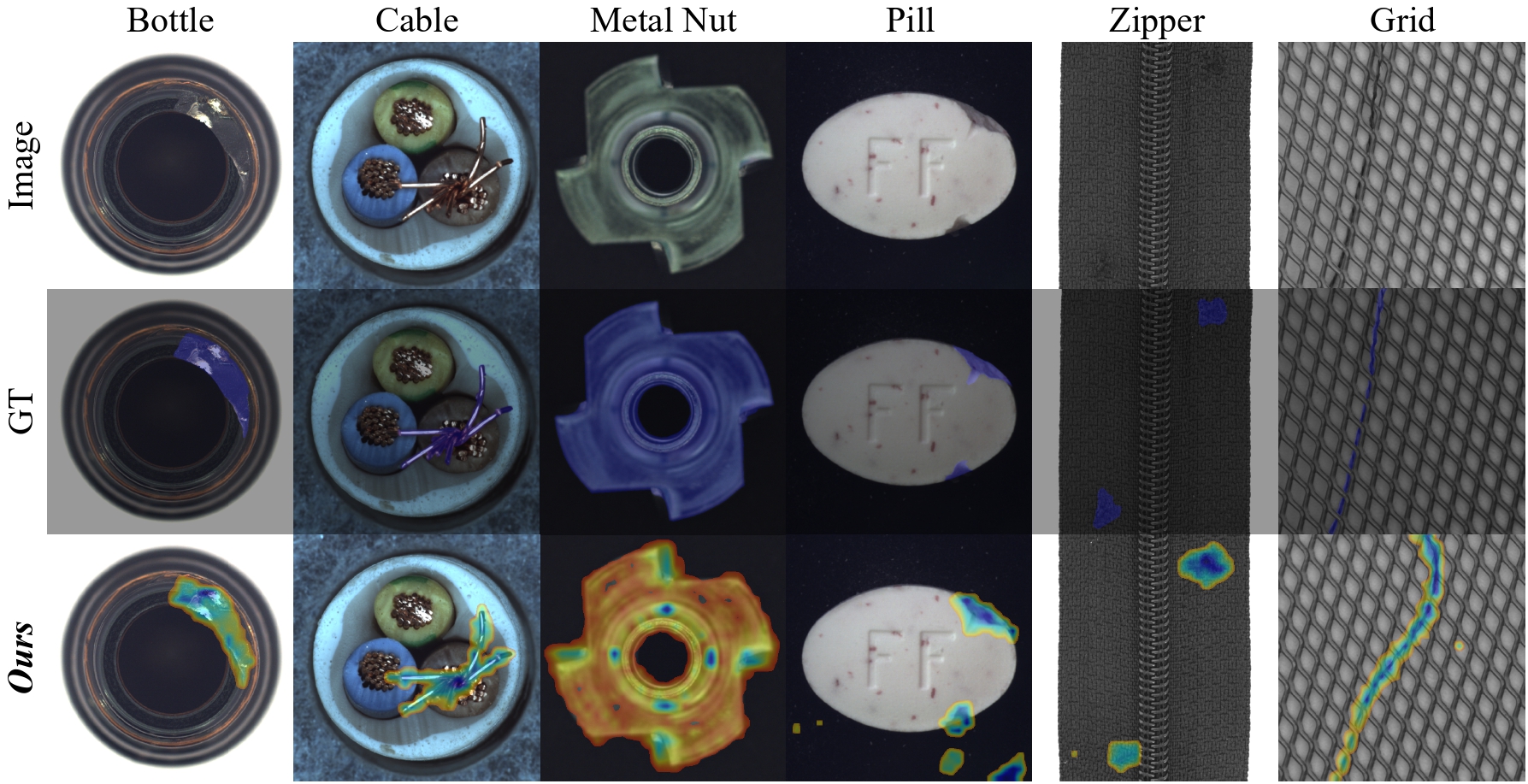}
    \caption{\textit{Additional qualitative results on MVTec-AD.}
    Examples from six categories (\textit{Bottle}, \textit{Cable}, \textit{Metal Nut}, \textit{Pill}, \textit{Zipper}, \textit{Grid}). 
    Rows show the input image, ground-truth mask, and our prediction. 
    \textit{SubspaceAD} effectively localizes structural, positional, and surface anomalies across varied MVTec categories.}
    \label{fig:mvtec_qualitative}
\end{figure}

To complement the qualitative experiments in the main paper (Fig.~\ref{fig:intro_figure} and Fig.~\ref{fig:qualitative_combined}), this section provides a more extensive set of qualitative results. We present detailed anomaly maps for representative samples from the VisA dataset in Fig.~\ref{fig:visA_qualitative} and the MVTec-AD dataset in Fig.~\ref{fig:mvtec_qualitative}. These examples further illustrate the model's localization performance across diverse object categories and anomaly types.

\section{Failure Cases and Limitations}
\label{sec:failure_cases}
\begin{figure}[ht]
    \centering
    \includegraphics[width=\linewidth]{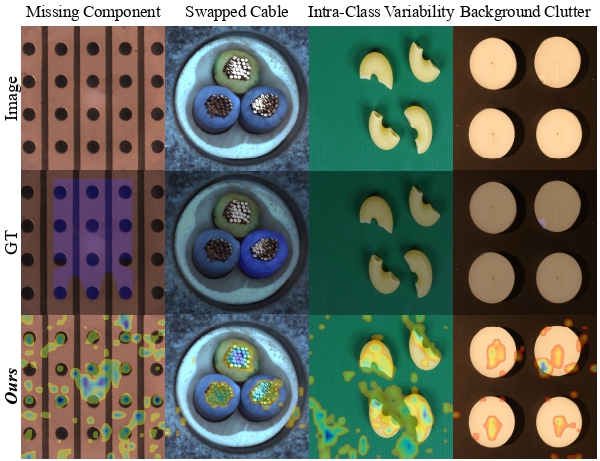}
    \caption{Qualitative failure cases. Each row shows the image, ground-truth mask, and our anomaly map. Examples include missed structural defects (e.g., missing transistor component), incorrect detection of cable swaps, and false positives from intra-class variability or background clutter.}
    \label{fig:failure_cases}
\end{figure}

While SubspaceAD demonstrates strong performance, it is important to acknowledge its limitations, particularly those inherent to patch-based, few-shot methodologies. Two primary modes of failure are identified, which are common challenges in this domain and are illustrated in Fig.~\ref{fig:failure_cases}.

\paragraph{Logical and Structural Anomalies}
As a patch-based method, SubspaceAD excels at modeling the \textit{local appearance} and \textit{texture} of normal samples. However, it does not explicitly model global spatial relationships or semantic rules. Consequently, it struggles with logical anomalies, such as a missing component. For example, in the MVTec-AD \textit{Transistor} category (discussed in Appendix~\ref{sec:per_category}), when a transistor is missing, the exposed circuit-board background may be incorrectly identified as normal texture. The model lacks the semantic, object-level understanding to know that a component \textit{should} be present in that specific location.

\paragraph{Background Artifacts and High Intra-Class Variance}
The model struggles in categories with high normal variance or complex, cluttered backgrounds, as seen in some parts of the VisA dataset. As noted in Appendix~\ref{sec:per_category}, categories Macaroni2 and PCB3 are challenging because their normal samples exhibit significant variation. This makes it difficult to form a single, compact subspace of normality from only a few shots. Furthermore, benign background artifacts (e.g., shadows, debris) that are not present in the few-shot support set may be incorrectly flagged as anomalies, since the model has no mechanism to infer that such regions belong to the normal background.

\paragraph{Outlook}
Despite these limitations, the overall results demonstrate that even a simple, training-free subspace model surpasses far more complex approaches in both detection accuracy and efficiency. The clarity of its statistical formulation, combined with its strong few-shot generalization, highlights the potential of foundation-model representations when paired with lightweight, interpretable modeling. Future work may extend this direction by incorporating geometric or semantic priors to better handle logical and structural anomalies.

\section{Impact of Rotation-Agnostic Preprocessing}
\label{sec:rotation_agnostic}
In the standard MVTec-AD dataset, the orientation of the \textit{Transistor} object is strictly fixed, meaning misrotation explicitly constitutes an anomaly. To assess the effect of a uniform, rotation-agnostic preprocessing pipeline across all categories, we conducted an ablation where random rotations were applied to the normal support samples.

However, for orientation-dependent categories like \textit{Transistor}, rotating the normal samples during PCA fitting fundamentally alters the model's definition of normality. By incorporating rotated features into the normal subspace, the model inadvertently learns to accept rotational defects as normal variations, causing it to miss actual misrotation anomalies during inference. 

Table~\ref{tab:transistor_rotation} outlines the performance under this rotation-agnostic protocol. As expected, forcing the subspace to account for rotational variance leads to explicit performance drops compared to the standard aligned baseline (cf. Table~\ref{tab:mvtec_detailed_full}). In the 1-shot setting, Image AUROC decreases by 7.5 percentage points, from 96.6\% to 89.1\%, while Pixel PRO drops by 6.3 percentage points, from 68.9\% to 62.6\%. This degradation persists in the 4-shot setting, where Pixel PRO decreases by 8.3 percentage points, from 72.7\% to 64.4\%. These results demonstrate that applying uniform rotational preprocessing is counterproductive for categories where orientation is a defining characteristic of the normal state.

\begin{table}[ht!]
\caption{Few-shot anomaly segmentation results of \textit{SubspaceAD} on the MVTec-AD \textit{Transistor} category using a \textbf{rotation-agnostic} protocol. We report mean Image AUROC (\%), Image AUPR (\%), Pixel AUROC (\%), and Pixel PRO (\%) $\pm$ standard deviation across 5 seeds.}
\label{tab:transistor_rotation}
\centering
\small
\begin{tabular}{@{}l c c c c@{}}
\toprule
\textbf{Shots} & \textbf{Img AUROC} & \textbf{Img AUPR} & \textbf{Pxl AUROC} & \textbf{Pxl PRO} \\ \midrule
1-shot & 89.1 $\pm$ 7.5 & 84.7 $\pm$ 5.7 & 82.3 $\pm$ 3.1 & 62.6 $\pm$ 2.0 \\
2-shot & 92.6 $\pm$ 3.6 & 88.5 $\pm$ 4.7 & 83.6 $\pm$ 0.9 & 63.7 $\pm$ 0.9 \\
4-shot & 93.4 $\pm$ 4.0 & 90.3 $\pm$ 4.3 & 84.0 $\pm$ 1.9 & 64.4 $\pm$ 1.3 \\
\bottomrule
\end{tabular}
\end{table}
\end{document}